\documentclass[sigconf]{acmart}
\AtBeginDocument{%
  \providecommand\BibTeX{{%
    \normalfont B\kern-0.5em{\scshape i\kern-0.25em b}\kern-0.8em\TeX}}}

\usepackage{hyperref}
\usepackage{url}
\usepackage{graphicx}
\usepackage{amsmath}
\usepackage{subcaption}
\usepackage{booktabs}
\usepackage{tablefootnote}
\usepackage{multirow}

\copyrightyear{2022} 
\acmYear{2022} 
\setcopyright{acmcopyright}
\acmConference[MM '22]{Proceedings of the 30th ACM International Conference on Multimedia}{October 10--14, 2022}{Lisboa, Portugal}
\acmBooktitle{Proceedings of the 30th ACM International Conference on Multimedia (MM '22), October 10--14, 2022, Lisboa, Portugal}
\acmPrice{15.00}
\acmDOI{10.1145/3503161.3547911}
\acmISBN{978-1-4503-9203-7/22/10}

\begin{document}
\fancyhead{}
\title{DiT: Self-supervised Pre-training for \\Document Image Transformer}


\author{Junlong Li}
\email{lockonn@sjtu.edu.cn}
\authornote{Contributions during internship at Microsoft Research Asia. Corresponding authors: Lei Cui and Furu Wei.}
\affiliation{
  \institution{Shanghai Jiao Tong University}
  \city{Shanghai}
  \country{China}
}

\author{Yiheng Xu}
\email{t-yihengxu@microsoft.com}
\authornotemark[1]
\affiliation{
  \institution{Microsoft Research Asia}
  \city{Beijing}
  \country{China}
}

\author{Tengchao Lv}
\email{tengchaolv@microsoft.com}
\affiliation{
  \institution{Microsoft Research Asia}
  \city{Beijing}
  \country{China}
}

\author{Lei Cui}
\email{lecu@microsoft.com}
\affiliation{
  \institution{Microsoft Research Asia}
  \city{Beijing}
  \country{China}
}

\author{Cha Zhang}
\email{chazhang@microsoft.com}
\affiliation{
  \institution{Microsoft Azure AI}
  \city{Redmond}
  \country{United States}
}

\author{Furu Wei}
\email{fuwei@microsoft.com}
\affiliation{
  \institution{Microsoft Research Asia}
  \city{Beijing}
  \country{China}
}

\renewcommand{\shortauthors}{Li et.al.}

\begin{abstract}
Image Transformer has recently achieved significant progress for natural image understanding, either using supervised (ViT, DeiT, etc.) or self-supervised (BEiT, MAE, etc.) pre-training techniques. In this paper, we propose \textbf{DiT}, a self-supervised pre-trained \textbf{D}ocument \textbf{I}mage \textbf{T}ransformer model using large-scale unlabeled text images for Document AI tasks, which is essential since no supervised counterparts ever exist due to the lack of human-labeled document images. We leverage DiT as the backbone network in a variety of vision-based Document AI tasks, including document image classification, document layout analysis, table detection as well as text detection for OCR. Experiment results have illustrated that the self-supervised pre-trained DiT model achieves new state-of-the-art results on these downstream tasks, e.g. document image classification (91.11 $\rightarrow$ 92.69), document layout analysis (91.0 $\rightarrow$ 94.9), table detection (94.23 $\rightarrow$ 96.55) and text detection for OCR (93.07 $\rightarrow$ 94.29). The code and pre-trained models are publicly available at \url{https://aka.ms/msdit}.
\end{abstract}





\maketitle

\section{Introduction}

Self-supervised pre-training techniques have been the de facto common practice for Document AI~\citep{cui2021document} in the past several years, where the image, text, and layout information is often jointly trained using a unified Transformer architecture~\citep{10.1145/3394486.3403172,xu2021layoutlmv2, huang2022layoutlmv3,xu2021layoutxlm,xu-etal-2022-xfund,pramanik2020multimodal,garncarek2021lambert,hong2021bros,powalski2021going,wu2021lampret,li2021structurallm,li2021selfdoc,appalaraju2021docformer}. Among all these approaches, a typical pipeline for pre-training Document AI models usually start with the vision-based understanding such as Optical Character Recognition (OCR) or document layout analysis, which still heavily relies on the supervised computer vision backbone models with human-labeled training samples. Although good results have been achieved on benchmark datasets, these vision models are often confronted with the performance gap in real-world applications due to domain shift and template/format mismatch from the training data. Such accuracy regression~\citep{li-etal-2020-tablebank,zhong2019publaynet} also has an essential influence on the pre-trained models as well as downstream tasks. Therefore, it is inevitable to investigate how to leverage the self-supervised pre-training for the backbone of document image understanding, which can better facilitate general Document AI models for different domains.

\begin{figure*}[t]
\centering
    \begin{subfigure}[b]{0.25\textwidth}
        \centering
        \fbox{\includegraphics[width=0.8\textwidth, height=1.1\textwidth]{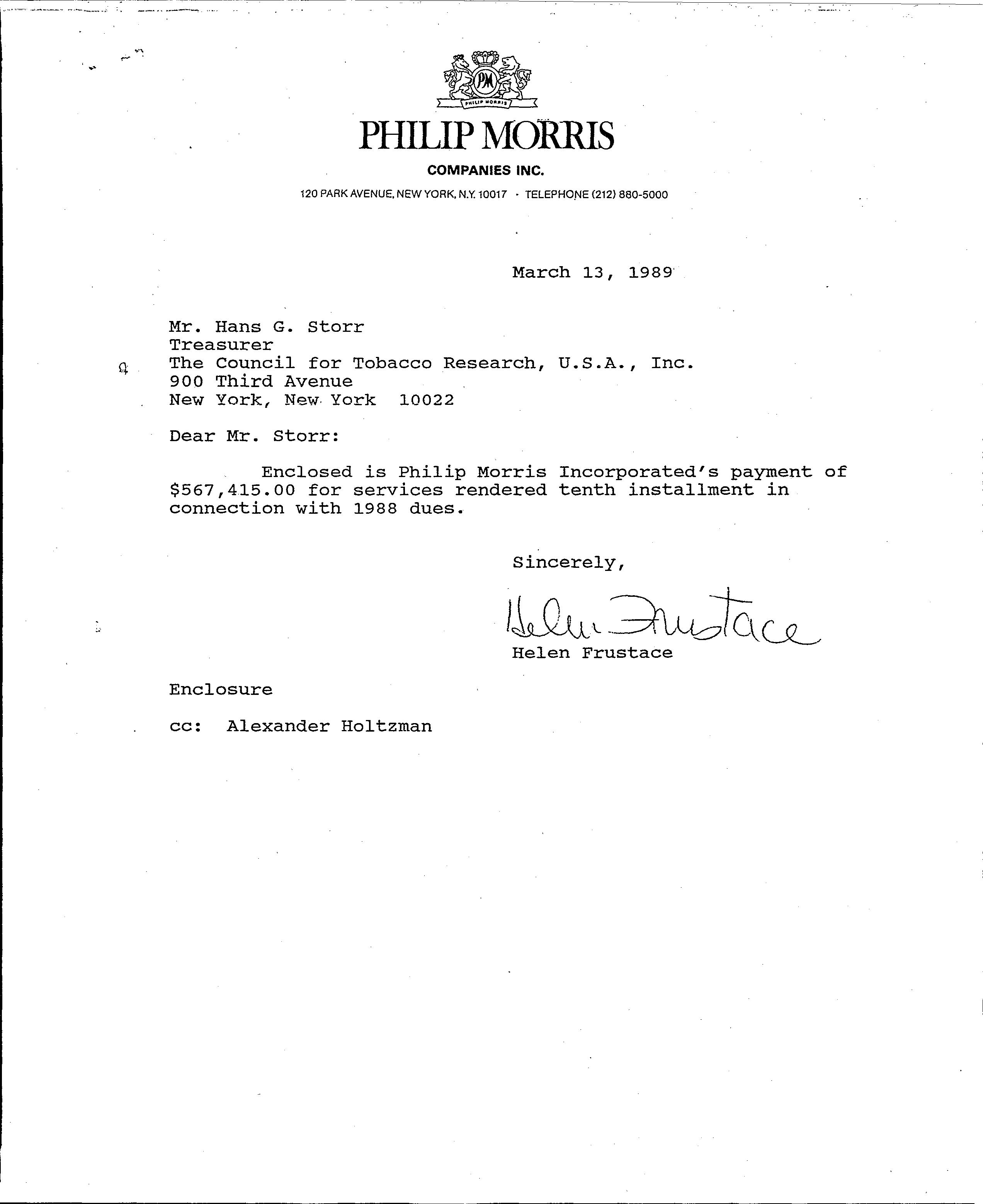}}
        \caption{}
        \label{fig:1a}
    \end{subfigure}
    ~ %add desired spacing between images, e. g. ~, \quad, \qquad, \hfill etc. 
    \begin{subfigure}[b]{0.25\textwidth}
        \centering
        \fbox{\includegraphics[width=0.8\textwidth, height=1.1\textwidth]{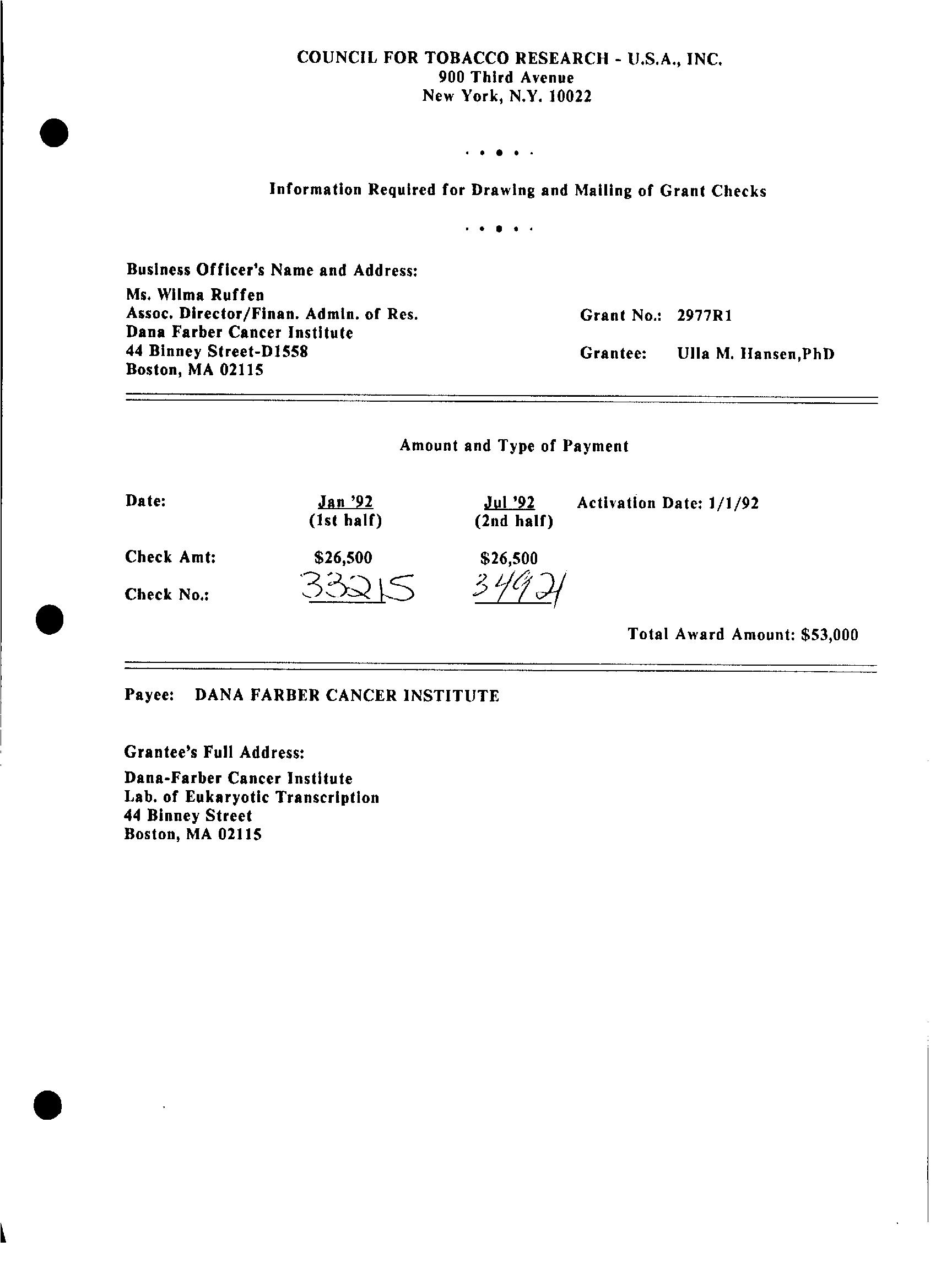}}
        \caption{}
        \label{fig:1b}
    \end{subfigure}
    ~ %add desired spacing between images, e. g. ~, \quad, \qquad, \hfill etc. 
    \begin{subfigure}[b]{0.25\textwidth}
        \centering
        \fbox{\includegraphics[width=0.8\textwidth, height=1.1\textwidth]{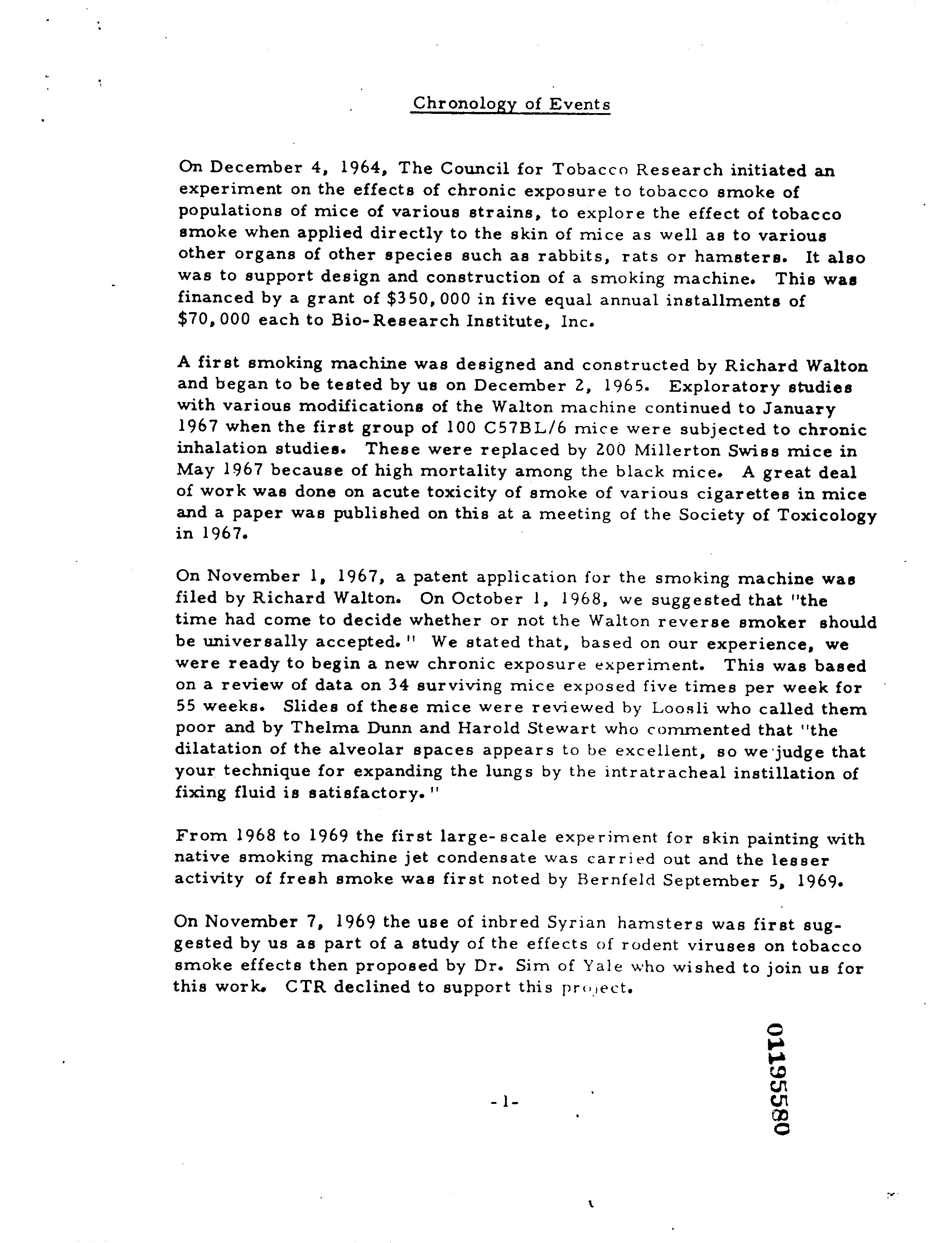}}
        \caption{}
        \label{fig:1c}
    \end{subfigure}
    ~
    \begin{subfigure}[b]{0.25\textwidth}
        \centering
        \fbox{\includegraphics[width=0.8\textwidth, height=1.1\textwidth]{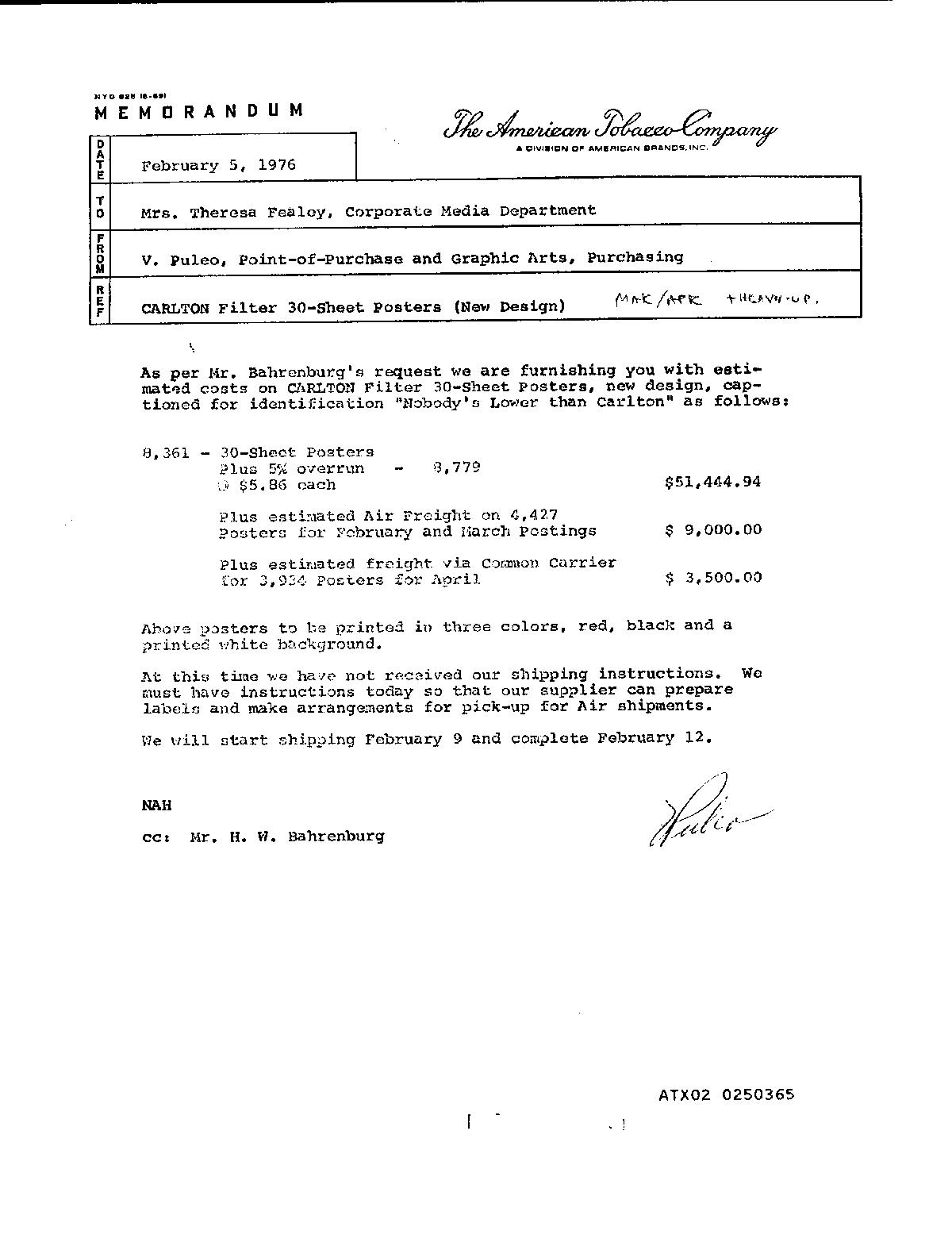}}
        \caption{}
        \label{fig:1d}
    \end{subfigure}
    ~
    \newline
    \begin{subfigure}[b]{0.25\textwidth}
        \centering
        \fbox{\includegraphics[width=0.8\textwidth, height=1.1\textwidth]{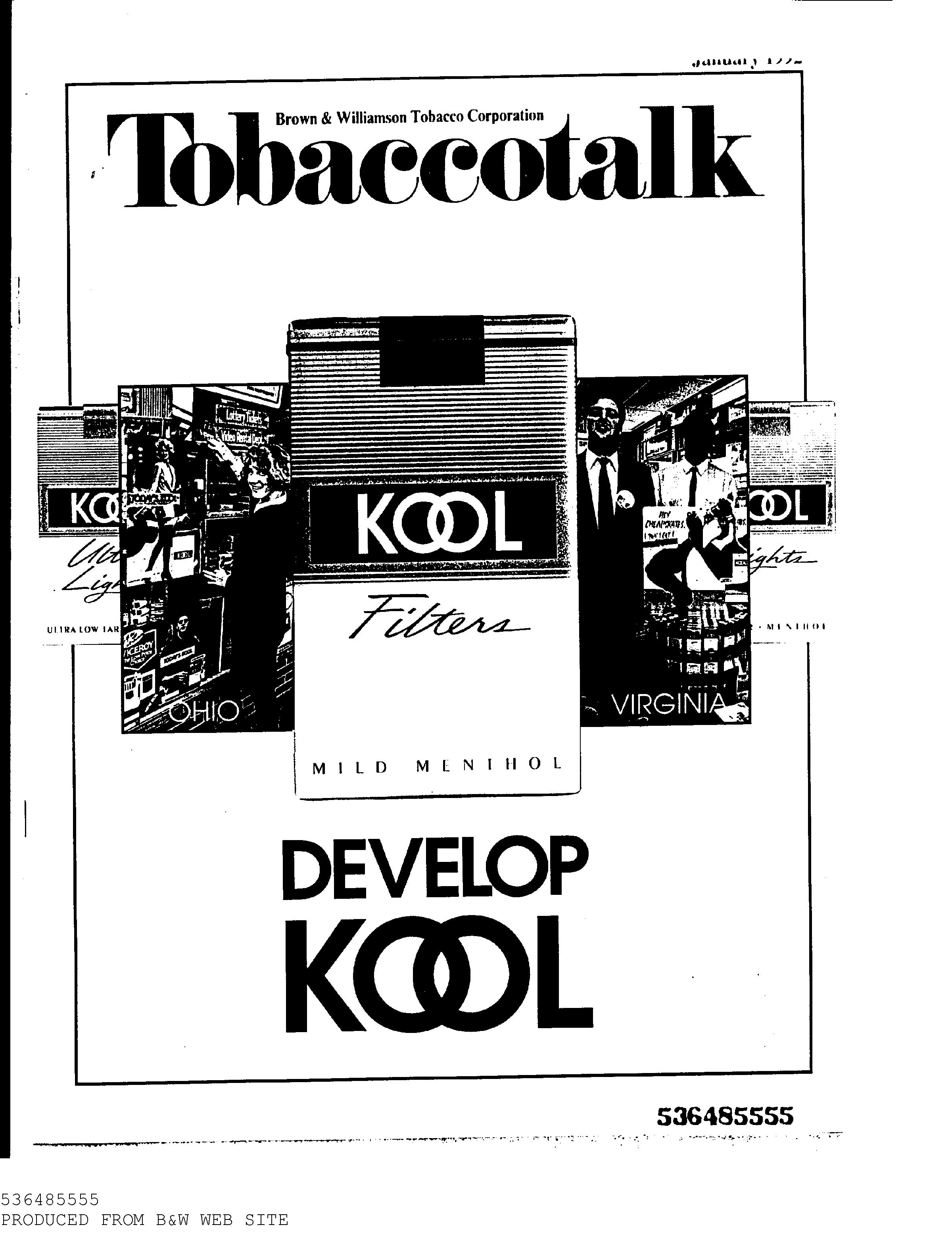}}
        \caption{}
        \label{fig:1e}
    \end{subfigure}
    ~
    \begin{subfigure}[b]{0.25\textwidth}
        \centering
        \fbox{\includegraphics[width=0.8\textwidth, height=1.1\textwidth]{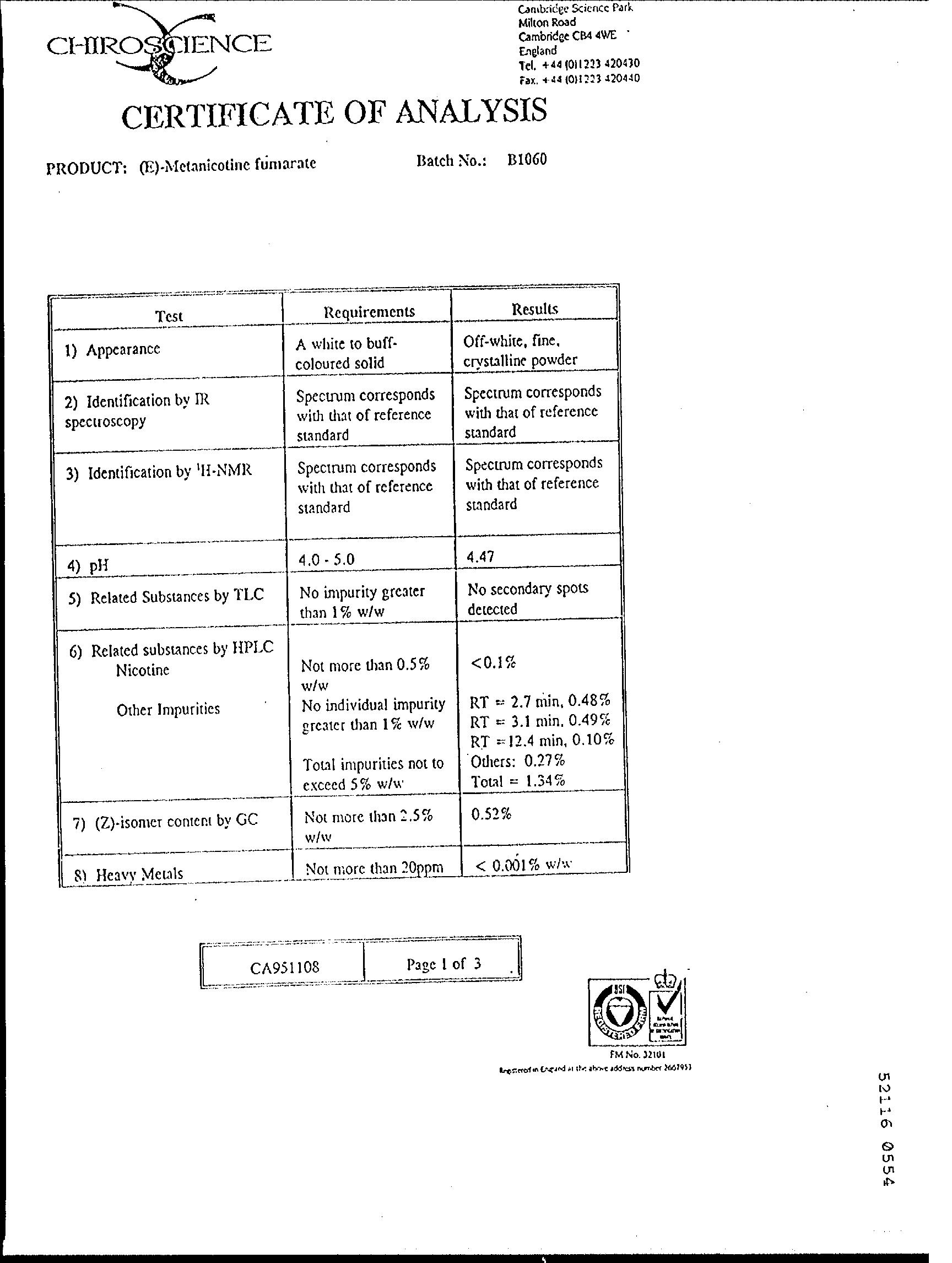}}
        \caption{}
        \label{fig:1f}
    \end{subfigure}
    ~
    \begin{subfigure}[b]{0.25\textwidth}
        \centering
        \fbox{\includegraphics[width=0.8\textwidth, height=1.1\textwidth]{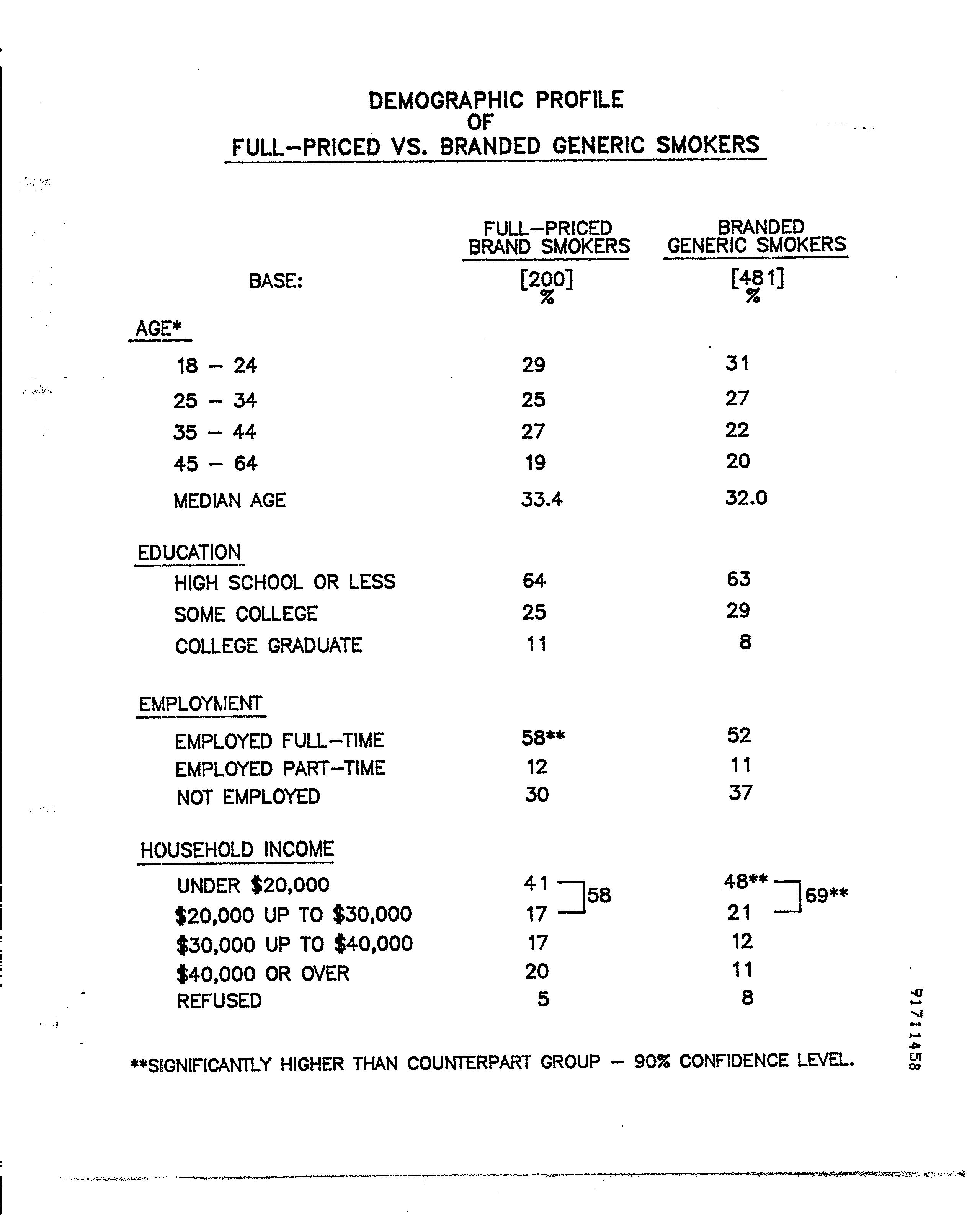}}
        \caption{}
        \label{fig:1g}
    \end{subfigure}
    ~
    \begin{subfigure}[b]{0.25\textwidth}
        \centering
        \fbox{\includegraphics[width=0.8\textwidth, height=1.1\textwidth]{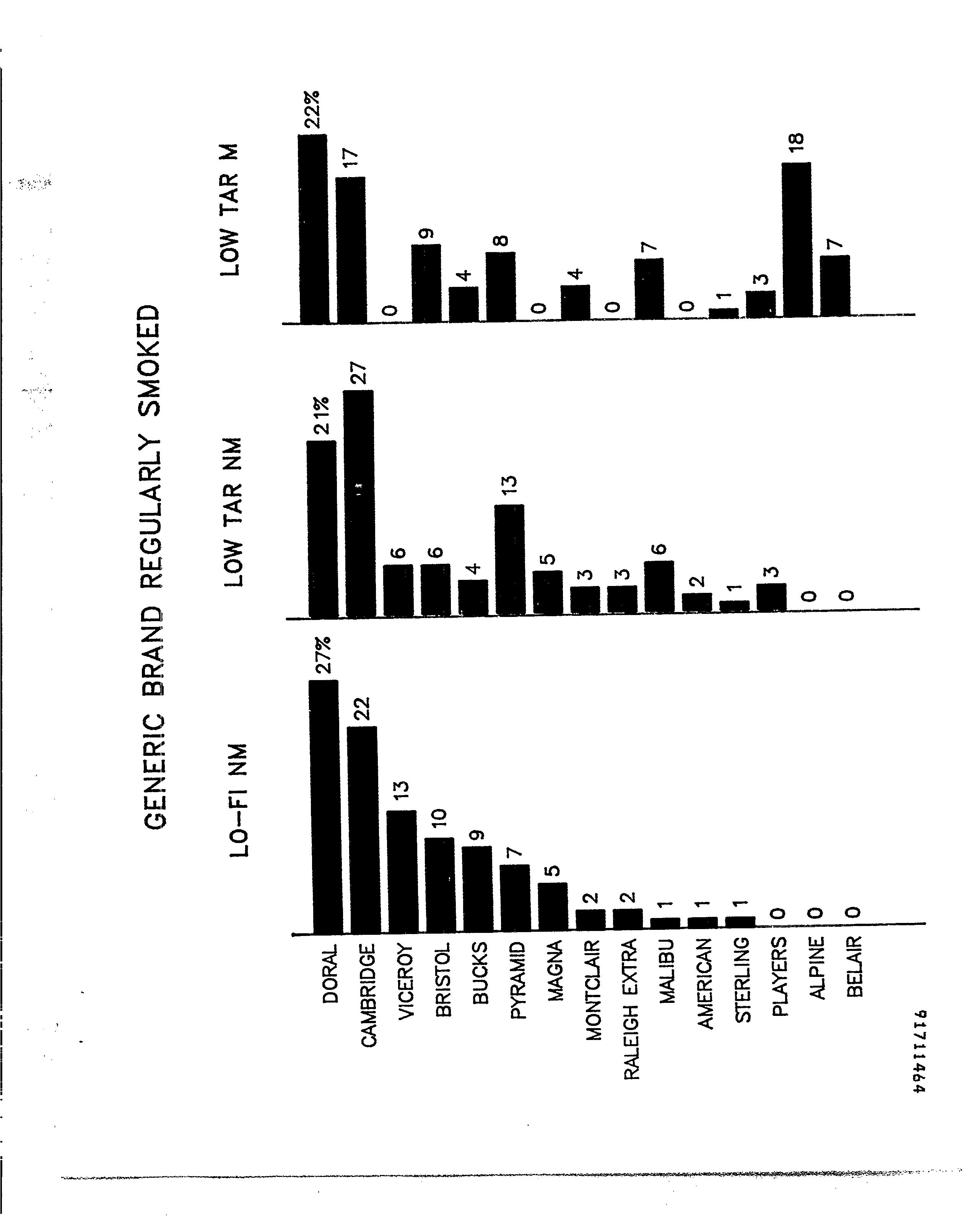}}
        \caption{}
        \label{fig:1h}
    \end{subfigure}
    \caption{Visually-rich business documents with different layouts and formats for pre-training DiT.}\label{fig:1}
\end{figure*}

Image Transformer~\citep{dosovitskiy2020vit,touvron2020deit,liu2021swin,chen2021empirical,bao2021beit,elnouby2021xcit,he2021masked,zhou2021ibot} has recently achieved great success for natural image understanding including classification, detection and segmentation tasks, either with supervised pre-training on the ImageNet or self-supervised pre-training. The pre-trained image Transformer models can achieve comparable and even better performance compared with CNN-based pre-trained models under a similar parameter size. However, for document image understanding, there is no commonly-used large-scale human-labeled benchmark like ImageNet, which makes large-scale supervised pre-training impractical. Even though weakly supervised methods have been used to create Document AI benchmarks~\citep{zhong2019publaynet,zhong2020imagebased,li-etal-2020-tablebank,li-etal-2020-docbank}, the domain of these datasets is often from the academic papers that share similar templates and formats, which are different from real-world documents such as forms, invoice/receipts, reports, and many others as shown in Figure~\ref{fig:1}. This may lead to unsatisfactory results for general Document AI problems. Therefore, it is vital to pre-train the document image backbone models with large-scale unlabeled data from general domains, which can support a variety of Document AI tasks.

To this end, we propose \textbf{DiT}, a self-supervised pre-trained Document Image Transformer model for general Document AI tasks, which does not rely on any human-labeled document images. Inspired by the recently proposed BEiT model~\citep{bao2021beit}, we adopt a similar pre-training strategy using document images. An input text image is first resized into $224\times224$ and then the image is split into a sequence of $16\times16$ patches which are used as the input to the image Transformer. Distinct from the BEiT model where visual tokens are from the discrete VAE in DALL-E~\citep{ramesh2021zeroshot}, we re-train the discrete VAE (dVAE) model with large-scale document images, so that the generated visual tokens are more domain relevant to the Document AI tasks. The pre-training objective is to recover visual tokens from dVAE based on the corrupted input document images using the Masked Image Modeling (MIM) in BEiT. In this way, the DiT model does not rely on any human-labeled document images, but only leverages large-scale unlabeled data to learn the global patch relationship within each document image. We evaluate the pre-trained DiT models on four publicly available Document AI benchmarks, including the RVL-CDIP dataset~\citep{harley2015icdar} for document image classification, the PubLayNet dataset~\citep{zhong2019publaynet} for document layout analysis, the ICDAR 2019 cTDaR dataset~\citep{8978120} for table detection, as well as the FUNSD dataset~\citep{Jaume2019FUNSDAD} for OCR text detection. Experiment results have illustrated that the pre-trained DiT model has outperformed the existing supervised and self-supervised pre-trained models and achieved new state-of-the-art on these tasks.

The contributions of this paper are summarized as follows:

\begin{enumerate}
    \item We propose DiT, a self-supervised pre-trained document image Transformer model, which can leverage large-scale unlabeled document images for pre-training.
    \item We leverage the pre-trained DiT models as the backbone for a variety of Document AI tasks, including document image classification, document layout analysis, table detection, as well as text detection for OCR, and achieve new state-of-the-art results.
    \item The code and pre-trained models are publicly available at \url{https://aka.ms/msdit}.
\end{enumerate}

\begin{figure*}[t]
    \centering
	\includegraphics[height=0.53\textwidth]{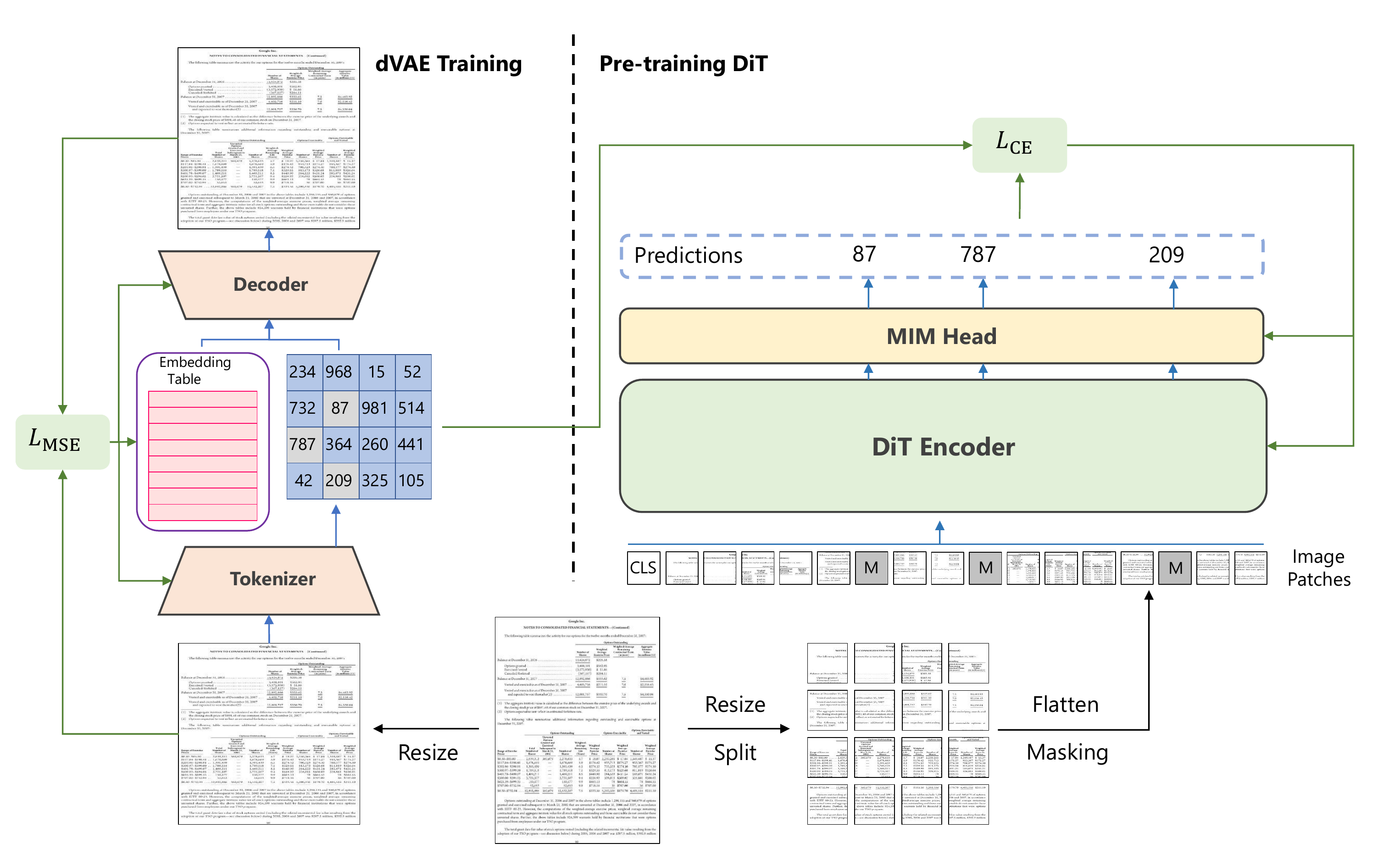}	
    \caption{The model architecture of DiT with MIM pre-training.}
    \label{arch}
\end{figure*}

\section{Related Work}

Image Transformer has recently achieved significant progress in computer vision problems, including classification, object detection, and segmentation.~\citep{dosovitskiy2020vit} first applied a standard Transformer directly to images with the fewest modifications. They split an image into $16\times16$ patches and provide the sequence of linear embeddings of these patches as an input to a Transformer named ViT. The ViT model is trained on image classification in a supervised fashion and outperforms the ResNet baselines. ~\citep{touvron2020deit} proposed data-efficient image transformers \& distillation through attention, namely DeiT, which solely relies on the ImageNet dataset for supervised pre-training and achieves SOTA results compared with ViT. ~\citep{liu2021swin} proposed a hierarchical Transformer whose
representation is computed with shifted windows. The shifted windowing scheme brings efficiency by limiting self-attention computation to non-overlapping local windows while also allowing for cross-window connection. In addition to supervised pre-trained models, ~\citep{pmlr-v119-chen20s} trained a sequence Transformer called iGPT to auto-regressively predict pixels without incorporating knowledge of the 2D input structure, which is the first attempt at self-supervised image transformer pre-training. After that, self-supervised pre-training for image Transformer became a hot topic in computer vision.~\citep{caron2021emerging} proposed DINO, which pre-trains the image Transformer using self-distillation with no labels.~\citep{chen2021empirical} proposed MoCov3 that is based on Siamese networks for self-supervised learning. More recently,~\citep{bao2021beit} adopted a BERT-style pre-training strategy, which first tokenizes the original image into visual tokens, then randomly masks some image patches and feeds them into the backbone Transformer. Similar to the masked language modeling, they proposed a masked image modeling task as the pre-training objective that achieves SOTA performance.~\citep{zhou2021ibot} presented a self-supervised framework iBOT that can perform masked prediction with an online tokenizer. The online tokenizer is jointly learnable with the MIM objective and dispenses with a multi-stage pipeline where the tokenizer is pre-trained beforehand.

The vision-based Document AI usually denote document analysis tasks that leverage the computer vision models, such as OCR, document layout analysis, and document image classification. Due to the lack of large-scale human-labeled datasets in this domain, existing approaches are usually based on the ConvNets models that are pre-trained with ImageNet/COCO datasets. Then, the models are continuously trained with task-specific labeled samples. To the best of our knowledge, the pre-trained DiT model is the first large-scale self-supervised pre-trained model for vision-based Document AI tasks. Meanwhile, it can be further leveraged for the multimodal pre-training for Document AI.

\section{Document Image Transformer}
In this section, we first present the architecture of DiT and the pre-training procedure. Then, we describe the application of DiT models in different downstream tasks.

\subsection{Model Architecture}
Following ViT~\citep{dosovitskiy2020vit}, we use the vanilla Transformer architecture~\citep{vaswani2017attention} as the backbone of DiT. We divide a document image into non-overlapping patches and obtain a sequence of patch embeddings. After adding the 1d position embedding, these image patches are passed into a stack of Transformer blocks with multi-head attention. Finally, we take the output of the Transformer encoder as the representation of image patches, which is shown in Figure~\ref{arch}.

\subsection{Pre-training}
Inspired by BEiT~\citep{bao2021beit}, we use Masked Image Modeling (MIM) as our pre-training objective. In this procedure, the images are represented as image patches and visual tokens in two views respectively. During pre-training, DiT accepts the image patches as input and predicts the visual tokens with the output representation. 

Like text tokens in natural language, an image can be represented as a sequence of discrete tokens obtained by an image tokenizer. BEiT uses the discrete variational auto-encoder (dVAE) from DALL-E~\citep{ramesh2021zeroshot} as the image tokenizer, which is trained on a large data collection including 400 million images. However, there exists a domain mismatch between natural images and document images, which makes the DALL-E tokenizer not appropriate for the document images. Therefore, to get better discrete visual tokens for the document image domain, we train a dVAE on the IIT-CDIP~\citep{Lewis:2006:BTC:1148170.1148307} dataset that includes 42 million document images.

To effectively pre-train the DiT model, we randomly mask a subset of inputs with a special token {\tt[MASK]} given a sequence of image patches. The DiT encoder embeds the masked patch sequence by a linear projection with added positional embeddings, and then contextualizes it with a stack of Transformer blocks. The model is required to predict the index of visual tokens with the output from masked positions. Instead of predicting the raw pixels, the masked image modeling task requires the model to predict the discrete visual tokens obtained by the image tokenizer.

\begin{figure}[t]
    \center
	\includegraphics[height=0.75\textwidth]{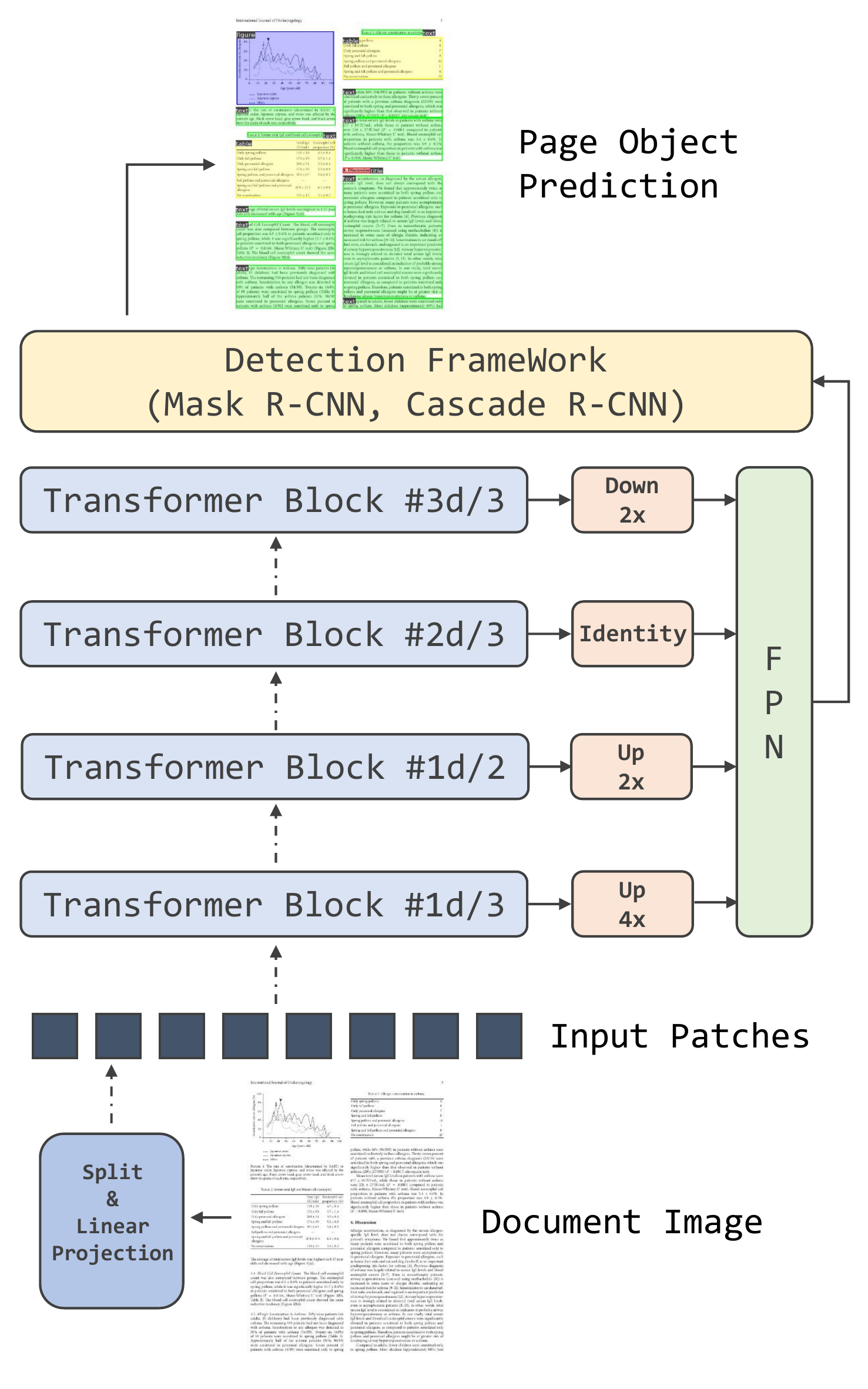}	
    \caption{Illustration of applying DiT as the backbone network in different detection frameworks.}
    \label{det}
\end{figure}

\begin{figure*}[t]
% 	\begin{subfigure}[b]{\textwidth}
%         \includegraphics[width=\textwidth]{tokenizer/cdip.jpg}
%         \caption{IIT-CDIP}
%         \label{fig:1h}
%     \end{subfigure}
%     ~
    % \newline
    \centering
    \begin{subfigure}[b]{1.0\textwidth}
        \centering
        \includegraphics[width=0.88\textwidth]{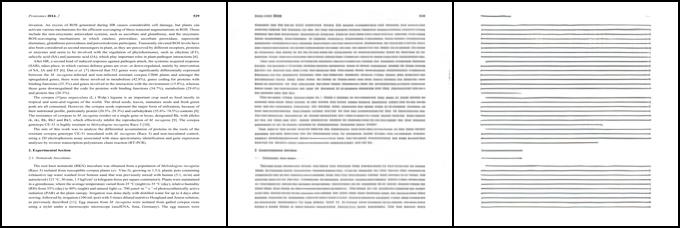}
        \caption{A sample from the PubLayNet dataset}
        \label{fig:tok1}
    \end{subfigure}
 
    \begin{subfigure}[b]{1.0\textwidth}
        \centering
        \includegraphics[width=0.88\textwidth]{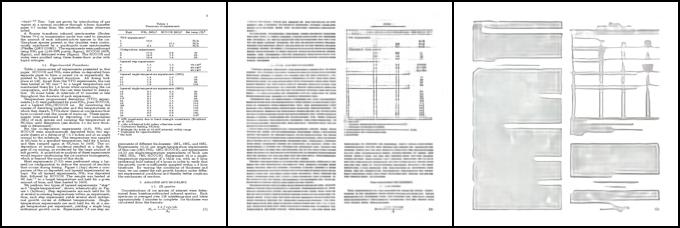}
        \caption{A sample from the ICDAR 2019 cTDaR dataset}
        \label{fig:tok2}
    \end{subfigure}
    \caption{Document image reconstruction with different tokenizers. From left to right: the original document image, image reconstruction using the self-trained dVAE tokenizer, image reconstruction using the DALL-E tokenizer.}\label{fig:tok}
\end{figure*}

\subsection{Fine-tuning}
We fine-tune our model on four Document AI benchmarks, including the RVL-CDIP dataset for document image classification, the PubLayNet dataset for document layout analysis, the ICDAR 2019 cTDaR dataset for table detection, and the FUNSD dataset for text detection. These benchmark datasets can be formalized as two common tasks: image classification and object detection.

\paragraph{Image Classification}
For image classification, we use average pooling to aggregate the representation of image patches. Next, we pass the global representation into a simple linear classifier.
 
\paragraph{Object Detection}
For object detection, as in Figure~\ref{det}, we leverage Mask R-CNN~\citep{he2017mask} and Cascade R-CNN~\citep{cai2018cascade} as detection frameworks and use ViT-based models as the backbone. Our code is implemented based on Detectron2~\citep{wu2019detectron2}. Following \citep{ElNouby2021XCiTCI, Li2021BenchmarkingDT}, we use  resolution-modifying modules at four different transformer blocks to adapt the single-scale ViT to the multi-scale FPN. Let $d$ be the total number of blocks, the $1d/3$th block is upsampled by 4$\times$ using a module with 2 stride-two 2$\times$2 transposed convolution.
% group normalization~\citep{wu2018group} and GeLU~\citep{hendrycks2016gaussian}
For the output of the $1d/2$th block, we use a single stride-two 2$\times$2 transposed convolution to upsample 2$\times$.
% (without normalization and non-linearity).
%The next $d/4$th block’s output is taken as it is and the final Transformer block’s output is downsampled by a factor of two using stride-two 2$\times$2 max pooling.
The output of the $2d/3$th block is utilized without additional operations.
Finally, the output of $3d/3$th block is downsampled by 2$\times$ with stride-two 2$\times$2 max pooling.
% Each of these modules preserves the embedding/channel dimension of the image Transformer. Assuming a patch size of $16\times16$, these modules produce feature maps with strides of 4, 8, 16, and 32 pixels, w.r.t. the input image, that is ready to be sent into an FPN. 

\section{Experiments}

\subsection{Tasks}
%We evaluate the performance of our model on four Document AI benchmarks, including the RVL-CDIP dataset for document image classification, the PubLayNet dataset for document layout analysis, and the ICDAR 2019 cTDaR dataset for table detection. 
We briefly introduce the datasets mentioned in section 3.3 here.

\paragraph{RVL-CDIP}
The RVL-CDIP~\citep{harley2015icdar} dataset consists of 400,000 grayscale images in 16 classes, with 25,000 per class. There are 320,000 training images, 40,000 validation images, and 40,000 test images. 
% The images are sized so their largest dimension does not exceed 1000 pixels.
The 16 classes include \{letter, form, email, handwritten, advertisement, scientific report, scientific publication, specification, file folder, news article, budget, invoice, presentation, questionnaire, resume, memo\}. The evaluation metric is the overall classification accuracy. 

\paragraph{PubLayNet}
 PubLayNet~\citep{zhong2019publaynet} is a large-scale document layout analysis dataset. More than 360,000 document images are constructed by automatically parsing PubMed XML files. The resulting annotations cover typical document layout elements such as text, title, list, figure, and table. The model needs to detect the regions of the assigned elements. We use the category-wise and overall mean average precision (MAP) @ intersection over union (IOU) [0.50:0.95] of bounding boxes as the evaluation metrics.

\paragraph{ICDAR 2019 cTDaR}
The cTDaR datasets~\citep{8978120} consist of two tracks, including table detection and table structure recognition. In this paper, we  focus on Track A where document images with one or several table annotations are provided. This dataset has two subsets, one for archival documents and the other for modern documents. The archival subset includes 600 training images and 199 testing images, which shows a wide variety of tables containing hand-drawn accounting books, stock exchange lists, train timetables, production census, etc. The modern subset consists of 600 training images and 240 testing images, which contain different kinds of PDF files, such as scientific journals, forms, financial statements, etc. The dataset contains Chinese and English documents in various formats, including scanned document images and born-digital formats. Metrics for evaluating this task are the precision, recall, and F1 scores computed from the model’s ranked output w.r.t. different Intersection over Union (IoU) threshold. We calculate the values with IoU thresholds of 0.6, 0.7, 0.8, and 0.9 respectively, and merge them into a final weighted F1 score: 

\begin{equation*}
wF1=\frac{0.6F1_{0.6}+0.7F1_{0.7}+0.8F1_{0.8}+0.9F1_{0.9}}{0.6+0.7+0.8+0.9}
\end{equation*}
This task further requires models to combine the modern and archival set as a whole to get a final evaluation result.

\paragraph{FUNSD}

FUNSD~\citep{Jaume2019FUNSDAD} is a noisy scanned document dataset labeled for three tasks: Text detection, Text recognition with Optical Character Recognition (OCR), and Form understanding. In this paper, we focus on Task \#1 in FUNSD, which aims to detect the text bounding boxes for scanned form documents. FUNSD includes 199 fully annotated forms with 31,485 words, whereas the training set contains 150 forms and the testing set includes 49 forms. The evaluation metrics are the precision, recall, and F1 score at IoU@0.5.

\subsection{Settings}

\paragraph{Pre-training Setup}
We pre-train DiT on the IIT-CDIP Test Collection 1.0~\citep{Lewis:2006:BTC:1148170.1148307}. We pre-process the dataset by splitting multi-page documents into single pages, and obtain 42 million document images. We also introduce random resized cropping to augment training data during training. We train our DiT-B model with the same architecture as the ViT base: a 12-layer Transformer with 768 hidden sizes, and 12 attention heads. The intermediate size of feed-forward networks is 3,072. A larger version, DiT-L, is also trained with 24 layers, 1,024 hidden sizes, and 16 attention heads. The intermediate size of feed-forward networks is 4,096.

\paragraph{The dVAE Tokenizer}
BEiT borrows the image tokenizer trained by DALL-E, which is not aligned with the document image data. In this case, we fully utilize the 42 million document images in the IIT-CDIP dataset and train a document dVAE image tokenizer to obtain the visual tokens. Like the DALL-E image tokenizer, the document image tokenizer has the codebook dimensionality of 8,192 and the image encoder with three layers. Each layer consists of a 2D convolution with a stride of 2 and a ResNet block. Therefore, the tokenizer eventually has a downsampling factor of 8. In this case, given a 112$\times$112 image, it ends up with a 14$\times$14 discrete token map aligning with the 14$\times$14 input patches.

We implement our dVAE codebase from open-sourced DALL-E implementation\footnote{\url{https://github.com/lucidrains/DALLE-pytorch}} and train the dVAE model with the entire IIT-CDIP dataset containing 42 million document images. The new dVAE tokenizer is trained with a combination of a MSE loss to reconstructe the input image, and a perplexity loss to increase the use of the quantized codebook representations. The input image size is 224$\times$224, and we train the tokenizer with a learning rate of 5e-4 and a minimum temperature of 1e-10 for 3 epochs.
We compare our dVAE tokenizer with the original DALL-E tokenizer by reconstructing the document image samples from downstream tasks, which is shown in Figure~\ref{fig:tok}. We sample images from the document layout analysis dataset PubLayNet and table detection dataset ICDAR 2019 cTDaR. After being reconstructed by the DALL-E and our tokenizer, the image tokenizer by DALL-E is hard to distinguish the border of lines and tokens, but the image tokenizer by our dVAE is closer to the original image and the border is sharper and clearer. We confirm that a better tokenizer can produce more accurate tokens that better describe the original images.

Equipped with the pre-training data and image tokenizer, we pre-train DiT for 500K steps with a batch size of 2,048, a learning rate of 1e-3, warmup steps of 10K, and weight decay of 0.05. The ${\beta}_{1}$ and ${\beta}_{2}$ of Adam~\citep{Kingma2015AdamAM} optimizer are 0.9 and 0.999 respectively. We employ stochastic depth~\citep{Huang2016DeepNW} with a 0.1 rate and disable dropout as in BEiT pre-training. We also apply blockwise masking in the pre-training of DiT with 40\% patches masked as BEiT.

\paragraph{Fine-tuning on RVL-CDIP}
We evaluate the pre-trained DiT models and other image backbones on RVL-CDIP for document image classification. We fine-tune the image transformers for 90 epochs with a batch size of 128 and a learning rate of 1e-3. For all settings, we resize the original images to 224 $\times$ 224 with the RandomResizedCrop operation.

\begin{figure}[t]
    \centering   
	\includegraphics[height=0.3\textwidth]{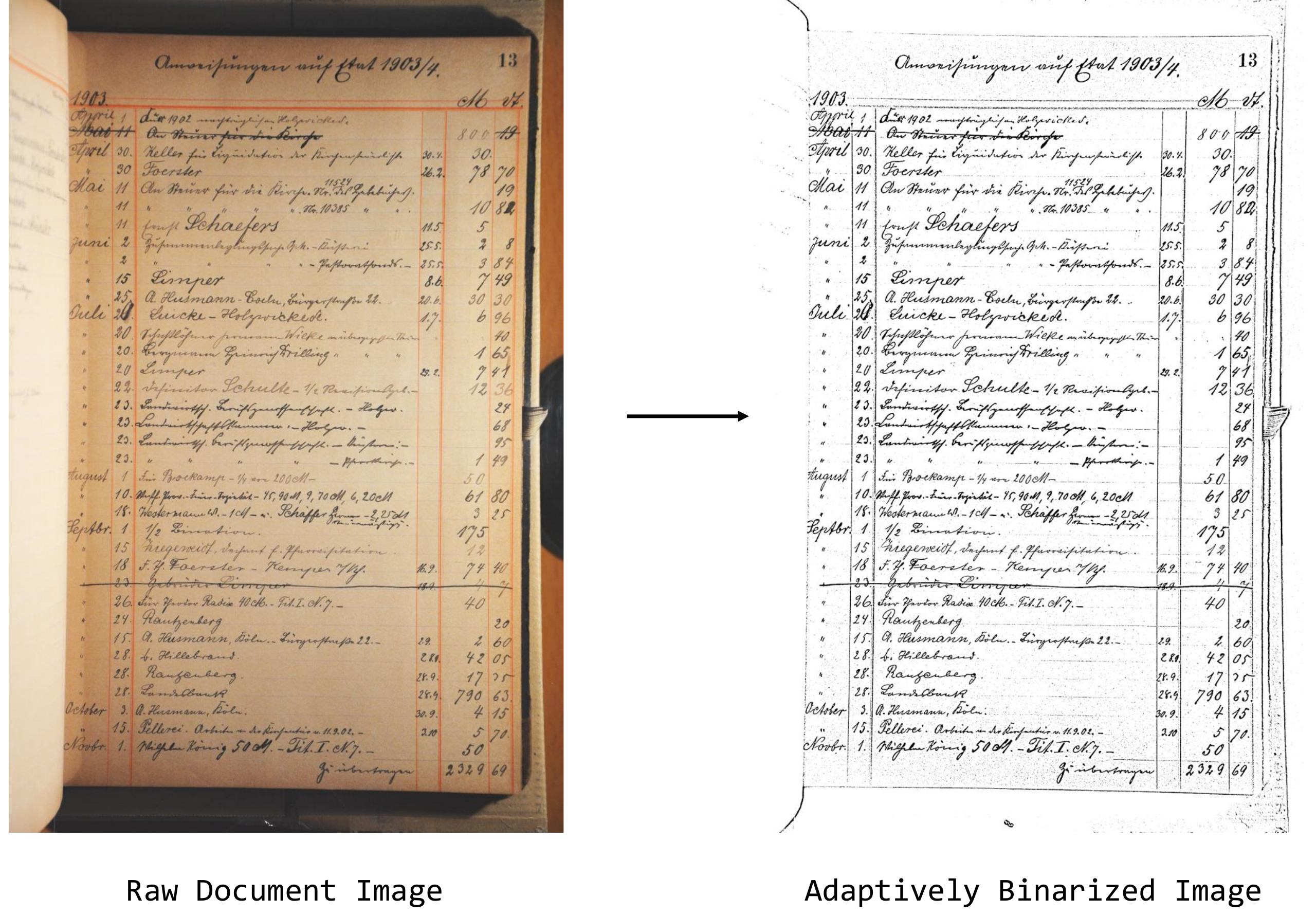}	
    \caption{An example of pre-processing with adaptive image binarization on the ICDAR 2019 cTDaR archival subset.}
    \label{bin}
\end{figure}

\begin{table}[t]
    \centering
    \begin{tabular}{cccc}
        \toprule
        \bf Model & \bf Type & \bf Accuracy & \bf \#Param \\
        \midrule
        \citep{Afzal2017CuttingTE}  & Single & 90.97 &  -\\
     \citep{Das2018DocumentIC}  & Single & 91.11 &- \\
     \citep{Das2018DocumentIC}  & Ensemble & 92.21 &- \\
    %  InceptionResNetV2\tablefootnote{\scriptsize \url{ https://medium.com/@jdegange85/benchmarking-modern-cnn-architectures-to-rvl-cdip-9dd0b7ec2955}}~\citep{Szegedy2016Inceptionv4IA} & 92.63\% & -\\
     \citep{ijcai2019-466}  & Ensemble & 92.77 &- \\
     \midrule
        
        ResNext-101-32$\times$8d & Single & 90.65 & 88M \\
        
        DeiT-B~\citep{touvron2020deit} & Single &  90.32 & 87M \\
        % DEiT-L \\
        BEiT-B~\citep{bao2021beit}  & Single &  91.09 & 87M\\
        MAE-B~\citep{he2021masked}  & Single &  91.42 & 87M \\
        % BEiT-L \\
        \midrule
        DiT-B   & Single & 92.11 & 87M \\
        DiT-L   & Single & \bf92.69 & 304M \\
        % DiT-L   \\
        
        \bottomrule
    \end{tabular}
    \caption{Document Image Classification accuracy (\%) on RVL-CDIP, where all the models use the pure image information (w/o text information) with the 224$\times$224 resolution.}
    \label{tab:rvlcdip}
\end{table}

\begin{table}[t]
    \centering
    \setlength{\tabcolsep}{3.0pt}
    \begin{tabular}{ccccccc}
        \toprule
        \multicolumn{1}{c}{\bf Model} & \bf Text & \bf Title & \bf List & \bf Table & \bf Figure & \bf Overall \\
        \midrule
        %Mask R-CNN \\
        \citep{zhong2019publaynet} & 0.916 & 0.840 & 0.886 & 0.960 & 0.949 & 0.910\\
        \midrule
        ResNext &0.916 &0.845 &0.918 &0.971 &0.952 &0.920 \\
        
        DeiT-B & 0.934 & 0.874 & 0.921 & 0.972 & 0.957 & 0.932\\
        % DEiT-L \\
        BEiT-B & 0.934 & 0.866 & 0.924 & 0.973 & 0.957 & 0.931\\
        MAE-B & 0.933 & 0.865 & 0.918 &  0.973 & 0.959 & 0.930 \\
        % BEiT-L \\
        \midrule
        DiT-B & 0.934 & 0.871 & 0.929 & 0.973 & 0.967 & 0.935   \\
        DiT-L & 0.937 & 0.879 & 0.945 & 0.974 & 0.968 & 0.941 \\
        \midrule
        ResNext (C) &0.930 &0.862 &0.940 &0.976 &0.968 &0.935\\
        DiT-B (C) & 0.944 & 0.889 & 0.948 & 0.976 & 0.969 & 0.945\\
        DiT-L (C) & \bf 0.944 & \bf 0.893 & \bf 0.960 & \bf 0.978 & \bf 0.972 & \bf 0.949\\
        %${\text{DiT}}_{R}$-B \\
        % DiT-L   \\
        \bottomrule
    \end{tabular}
    \caption{Document Layout Analysis mAP @ IOU [0.50:0.95] on PubLayNet validation set. ResNext-101-32$\times$8d is shortened as ResNext and Cascade as C.}
    \label{tab:layout}
\end{table}

%\vspace{-0.5cm}

\begin{table*}[t]
\begin{subtable}[ht]{1\textwidth}
    \centering
    \begin{tabular}{cccccc}
        \toprule

        \textbf{Model} & \textbf{IoU@0.6} & \textbf{IoU@0.7} & \textbf{IoU@0.8} & \textbf{IoU@0.9} & \textbf{WAvg. F1}\\
        \midrule
        1st place in cTDaR & 96.97 & 95.99 & 95.14 & 90.22 & 94.23 \\
        \midrule
        ResNeXt-101-32$\times$8d & 96.42 &95.99 & 95.15 &91.36 & 94.46\\
        DeiT-B &96.26	&95.56	&94.57	&90.91	&94.04\\
        BEiT-B &96.82	&96.40	&95.41	&92.44	&95.03\\
        MAE-B &96.86	&96.31	&95.05	&91.57	&94.66\\
        \midrule
        DiT-B &96.75 &96.19 &95.62 &93.36 &95.30\\
        DiT-L & \bf 97.83 & \bf 97.41 &96.29 &92.93 &95.85\\
        \midrule
        ResNeXt-101-32$\times$8d (Cascade) &96.54 &95.84 &95.13 &92.87 &94.90\\
        DiT-B (Cascade) &97.20 &96.92 &96.78 &94.26 &96.14\\
        DiT-L (Cascade)	&97.68	&97.26	& \bf 97.12	& \bf 94.74	& \bf 96.55 \\
        \bottomrule
    \end{tabular}
    \caption{Table detection accuracy on ICDAR 2019 cTDaR (combined: archival+modern)}
    \label{tab:table_combined}
\end{subtable}

\begin{subtable}[ht]{1\textwidth}
  \centering
    \begin{tabular}{cccccc}
        \toprule
        \textbf{Model} & \textbf{IoU@0.6} & \textbf{IoU@0.7} & \textbf{IoU@0.8} & \textbf{IoU@0.9} & \textbf{WAvg. F1}\\
        \midrule
        1st place in cTDaR  & 97.16  & 96.41  & 95.27  & 91.12 & 94.67 \\
        \midrule
        ResNeXt-101-32$\times$8d &96.60 &96.60 &95.09 &91.70 &94.73\\
        DeiT-B &97.54	&97.16	&96.41	&92.63	&95.68\\
        BEiT-B 	& \bf 98.10		& \bf 98.10		&95.82		&94.30	&96.35\\
        MAE-B 	&97.54		&97.54		&96.03		&94.14	&96.12\\
        \midrule
        DiT-B  &97.53 &97.15  &96.02 &94.88 &96.24\\
        DiT-L  &97.53  &97.15  &96.39  &95.26 &96.46 \\
        \midrule
        ResNeXt-101-32$\times$8d (Cascade)  &96.76 &96.38  &95.24 &93.71 &95.35\\
        DiT-B (Cascade)  &96.97  &96.97  &96.97  &95.83 &96.63 \\
        DiT-L (Cascade)	&97.34	&97.34	& \bf 97.34	& \bf 96.20	& \bf 97.00 \\
        \bottomrule
    \end{tabular}
    \caption{Table detection accuracy on ICDAR 2019 cTDaR (archival)}
    \label{tab:table_archival}
\end{subtable}

\begin{subtable}[ht]{1\textwidth}
    \centering
    \begin{tabular}{cccccc}
    \toprule
    \textbf{Model} & \textbf{IoU@0.6} & \textbf{IoU@0.7} & \textbf{IoU@0.8} & \textbf{IoU@0.9} & \textbf{WAvg. F1}\\
    \midrule
    1st place in cTDaR    &96.86 &95.74 &95.07  &89.69  &93.97 \\
    \midrule
    ResNeXt-101-32$\times$8d &96.30 &95.63  &95.18 &91.15 &94.30\\
    DeiT-B  & 95.51  & 94.61  & 93.48  & 89.89 & 93.07 \\
    BEiT-B  & 96.06  & 95.39  & 95.16  & 91.34 & 94.25 \\
    MAE-B  & 96.47  & 95.58  & 94.48  & 90.07 & 93.81 \\
    \midrule
    DiT-B 	&96.29 	&95.61		&95.39		&92.46	&94.74  \\
    DiT-L  &\bf 98.00 &\bf 97.56 &96.23 &91.57 &95.50 \\
    \midrule
    ResNeXt-101-32$\times$8d (Cascade) &96.41  &95.52 &95.07  &92.38 &94.63\\
    DiT-B (Cascade) 	&  97.33 &  96.89	&96.67	&93.33 &95.85 \\
    DiT-L (Cascade)		&97.89	&97.22	& \bf 97.00	& \bf 93.88	& \bf 96.29 \\

    \bottomrule
    \end{tabular}%
    \caption{Table detection accuracy on ICDAR 2019 cTDaR (modern)}
    \label{tab:table_modern}
\end{subtable}    
\caption{Table detection accuracy (F1) on ICDAR 2019 cTDaR.}
\label{tab:table_icdar}
\end{table*}

\paragraph{Fine-tuning on ICDAR 2019 cTDaR}
We evaluate the pre-trained DiT models and other image backbones on the ICDAR 2019 dataset for table detection. Since the image resolution for object detection tasks is much larger than classification, we limit the batch size to 16. The learning rate is 1e-4 and 5e-5 for archival and modern subsets respectively. In the preliminary experiments, we found that directly using the raw images in the archival subset leads to suboptimal performance when fine-tuning DiT, so we apply an adaptive image binarization algorithm implemented by OpenCV~\citep{opencv_library} to binarize the images. An example of the pre-procession is shown in Figure \ref{bin}. During training, we apply the data augmentation method used in DETR~\citep{carion2020end} as a multi-scale training strategy. Specifically, the input image is cropped with probability 0.5 to a random rectangular patch which is then resized again such that the shortest side is at least 480 and at most 800 pixels while the longest at most 1,333.

\paragraph{Fine-tuning on PubLayNet}
We evaluate the pre-trained DiT models and other image backbones on the PubLayNet dataset for document layout analysis. Similar to the ICDAR 2019 cTDaR dataset, the batch size is 16, and the learning rate is 4e-4 for the base version. and 1e-4 for the large version. The data augmentation method for DETR~\citep{carion2020end} is also used.

\paragraph{Fine-tuning on FUNSD}
We use the same object detection framework for fine-tuning the pre-trained DiT models and other backbones on the text detection task in FUNSD. In document layout analysis and table detection, we use anchor box sizes [32, 64, 128, 256, 512] in the detection process since the detected areas are usually paragraph-level. Different from document layout analysis, text detection aims to locate smaller objects at the word level in document images. Therefore, we use anchor box sizes [4, 8, 16, 32, 64] in the detection process. The batch size is set to 16 and the learning rate is 1e-4 for the base model and 5e-5 for the large model.

The image backbone models selected as baselines have a comparable number of parameters compared with our DiT-B. They include the following two kinds: CNN and image Transformer. For CNN-based models, we choose ResNext101-32$\times$8d~\citep{Xie2016AggregatedRT}. For image Transformers, we choose the base version of DeiT~\citep{touvron2020deit}, BEiT~\citep{bao2021beit} and MAE~\citep{he2021masked} which are pre-trained on ImageNet-1K dataset with a 224$\times$224 input size. We rerun the fine-tuning of all baselines.

\subsection{Results}

\paragraph{RVL-CDIP}
The results of document image classification on RVL-CDIP are shown in Table \ref{tab:rvlcdip}. To make a fair comparison, the approaches in the table use only image information from the dataset. DiT-B performs significantly better than all selected single-model baselines. Since DiT shares the same model structure with other image Transformer baselines, the higher score indicates the effectiveness of our document-specific pre-training strategy. The larger version, DiT-L, gets a comparable score with the previous SOTA ensemble model under the single-model setting, which further highlights its modeling capability on document images.

\paragraph{PubLayNet}
The results of document layout analysis on PubLayNet are shown in Table \ref{tab:layout}. Since this task has a large number of training and testing samples and requires a comprehensive analysis of the common document elements, it clearly demonstrates the learning ability of different image Transformer models. It is observed that the DeiT-B, BEiT-B, and MAE-B are obviously better than ResNeXt-101, and DiT-B is even stronger than these powerful image Transformer baselines. According to the results, the improvement mainly comes from the List and Figure category, and on the basis of DiT-B, DiT-L gives out a much higher mAP score. We also investigate the impact of different object detection algorithms, and the results show that a more advanced detection algorithm (Cascade R-CNN in our case) can push the model performance to a higher level. We also apply Cascade R-CNN on the ResNeXt-101-32$\times$8d baseline, and DiT surpasses it by 1\% and 1.4\% absolute score for the base and large settings respectively, indicating the superiority of DiT on a different detection framework.

\paragraph{ICDAR 2019 cTDaR}
The results of table detection on ICDAR 2019 cTDaR dataset are shown in Table \ref{tab:table_icdar}. The size of this dataset is relatively small, so it aims at evaluating the few-shot learning capability of models under a low-resource scenario. We first analyze the model performance on the archival and modern subsets separately. In Table~\ref{tab:table_archival}, DiT surpasses all the baselines except BEiT for the archival subset. 
This is because in the pre-training of BEiT, it directly uses the DALL-E dVAE which is trained on an extremely large dataset with 400M images with different colors. While for DiT, the image tokenizer is trained with grayscale images, which may not be sufficient for historical document images with colors. The improvement when switching from Mask R-CNN to Cascade R-CNN is also observed which is similar to PubLayNet settings, and DiT still outperforms other baselines significantly. The conclusion is similar to the results on the modern subset in Table \ref{tab:table_modern}. We further combine the predictions of the two subsets into a single set. The results in \ref{tab:table_combined} show DiT-L achieves the highest wF1 score among all Mask R-CNN methods, demonstrating the versatility of DiT under different categories of documents. It is worth noting that the metrics of IoU@0.9 are significantly better, which means DiT has a better fine-grained object detection capability. Under all the three settings, we have pushed the SOTA results to a new level by more than 2\% (94.23$\rightarrow$96.55) absolute wF1 score with our best model and the Cascade R-CNN detection algorithm. 

\begin{table}[t]
    \centering
    \begin{tabular}{cccc}
        \toprule
        \multicolumn{1}{c}{\bf Model}  & \bf Precision & \bf Recall & \bf F1  \\
        \midrule
        Faster R-CNN~\citep{Jaume2019FUNSDAD} & 0.704 & 0.848 & 0.76 \\
        DBNet~\citep{liao2022realtime} &0.8764 &0.8400 &0.8578 \\
        A Commercial OCR Engine &0.8762 &0.8260 &0.8504\\
        \midrule
        ResNeXt-101-32$\times$8d &0.9387 &0.9229 &0.9307 \\
        DeiT-B &0.9429 &0.9237 &0.9332\\
        BEiT-B &0.9412 &0.9263 &0.9337\\
        MAE-B &0.9441 &0.9321 &0.9381\\
        \midrule
        DiT-B &0.9470 &0.9307 &0.9388\\
        DiT-L &0.9452 & \bf 0.9336 &0.9393\\
        \midrule
        DiT-B (+syn) &0.9539 &0.9315 &0.9425\\
        DiT-L (+syn) &\bf 0.9543 &0.9317 & \bf 0.9429\\
        % \midrule
        % ResNeXt-101-32$\times$8d (Cascade) \\
        % DiT-B (Cascade)\\
        % DiT-L (Cascade)\\
        \bottomrule
    \end{tabular}
    \caption{Text detection accuracy (IoU@0.5) on FUNSD Task \#1, where Mask R-CNN is used with different backbones (ResNeXt, DeiT, BEiT, MAE and DiT). ``+syn'' denotes that DiT is trained with a synthetic dataset including 1M document images, then fine-tuned with the FUNSD training data.}
    \label{tab:funsd}
\end{table}

\paragraph{FUNSD (Text Detection)}

The results of text detection on the FUNSD dataset are shown in Table~\ref{tab:funsd}. Since text detection for OCR has been a long-standing real-world problem, we obtain the word-level text detection results from a popular commercial OCR engine to set a high-level baseline. In addition, DBNet~\citep{liao2022realtime} is a widely used text detection model for online OCR engines, we also fine-tune a pre-trained DBNet model with FUNSD training data and evaluate its accuracy. Both of them achieve around 0.85 F1 scores for IoU@0.5. Next, we use the Mask R-CNN framework to compare different backbone networks (CNN and ViT) including ResNeXt-101, DeiT, BEiT, MAE, and DiT. It is shown that CNN-based and ViT-based text detection models outperform the baselines significantly due to advanced model design and more parameters. We also observe that the DiT models achieve new SOTA results compared with other models. Finally, we further train the DiT models with a synthetic dataset that contains 1 million document images, leading to an F1 of 0.9429 being achieved by the DiT-L model.

\section{Conclusion and Future Work}

In this paper, we present DiT, a self-supervised foundation model for general Document AI tasks. The DiT model is pre-trained with large-scale unlabeled document images that cover a variety of templates and formats, which is ideal for downstream Document AI tasks in different domains. We evaluate the pre-trained DiT on several vision-based Document AI benchmarks, including  table detection, document layout analysis, document image classification, and text detection. Experimental results have shown that DiT outperforms several strong baselines across the board and achieves new SOTA performance. We will make the pre-trained DiT models publicly available to facilitate the Document AI research.

For future research, we will pre-train DiT with a much larger dataset to further push the SOTA results in Document AI. Meanwhile, we will also integrate DiT as the foundation model in multi-modal pre-training for visually-rich document understanding such as the next-gen layout-based models like LayoutLM, where a unified Transformer-based architecture may be sufficient for both CV and NLP applications in Document AI.

\bibliographystyle{ACM-Reference-Format}
\bibliography{base}


\begin{thebibliography}{48}

%%% the Web address.

\ifx \showCODEN    \undefined \def \showCODEN     #1{\unskip}     \fi
\ifx \showDOI      \undefined \def \showDOI       #1{#1}\fi
\ifx \showISBNx    \undefined \def \showISBNx     #1{\unskip}     \fi
\ifx \showISBNxiii \undefined \def \showISBNxiii  #1{\unskip}     \fi
\ifx \showISSN     \undefined \def \showISSN      #1{\unskip}     \fi
\ifx \showLCCN     \undefined \def \showLCCN      #1{\unskip}     \fi
\ifx \shownote     \undefined \def \shownote      #1{#1}          \fi
\ifx \showarticletitle \undefined \def \showarticletitle #1{#1}   \fi
\ifx \showURL      \undefined \def \showURL       {\relax}        \fi
% invisible to TeX
\providecommand\bibfield[2]{#2}
\providecommand\bibinfo[2]{#2}
\providecommand\natexlab[1]{#1}
\providecommand\showeprint[2][]{arXiv:#2}

\bibitem[Afzal et~al\mbox{.}(2017)]%
        {Afzal2017CuttingTE}
\bibfield{author}{\bibinfo{person}{Muhammad~Zeshan Afzal},
  \bibinfo{person}{Andreas K{\"o}lsch}, \bibinfo{person}{Sheraz Ahmed}, {and}
  \bibinfo{person}{Marcus Liwicki}.} \bibinfo{year}{2017}\natexlab{}.
\newblock \showarticletitle{Cutting the Error by Half: Investigation of Very
  Deep CNN and Advanced Training Strategies for Document Image Classification}.
\newblock \bibinfo{journal}{\emph{2017 14th IAPR International Conference on
  Document Analysis and Recognition (ICDAR)}}  \bibinfo{volume}{01}
  (\bibinfo{year}{2017}), \bibinfo{pages}{883--888}.
\newblock


\bibitem[Appalaraju et~al\mbox{.}(2021)]%
        {appalaraju2021docformer}
\bibfield{author}{\bibinfo{person}{Srikar Appalaraju}, \bibinfo{person}{Bhavan
  Jasani}, \bibinfo{person}{Bhargava~Urala Kota}, \bibinfo{person}{Yusheng
  Xie}, {and} \bibinfo{person}{R. Manmatha}.} \bibinfo{year}{2021}\natexlab{}.
\newblock \bibinfo{title}{DocFormer: End-to-End Transformer for Document
  Understanding}.
\newblock
\newblock
\showeprint[arxiv]{2106.11539}~[cs.CV]


\bibitem[Bao et~al\mbox{.}(2021)]%
        {bao2021beit}
\bibfield{author}{\bibinfo{person}{Hangbo Bao}, \bibinfo{person}{Li Dong},
  {and} \bibinfo{person}{Furu Wei}.} \bibinfo{year}{2021}\natexlab{}.
\newblock \bibinfo{title}{BEiT: BERT Pre-Training of Image Transformers}.
\newblock
\newblock
\showeprint[arxiv]{2106.08254}~[cs.CV]


\bibitem[Bradski(2000)]%
        {opencv_library}
\bibfield{author}{\bibinfo{person}{G. Bradski}.}
  \bibinfo{year}{2000}\natexlab{}.
\newblock \showarticletitle{{The OpenCV Library}}.
\newblock \bibinfo{journal}{\emph{Dr. Dobb's Journal of Software Tools}}
  (\bibinfo{year}{2000}).
\newblock


\bibitem[Cai and Vasconcelos(2018)]%
        {cai2018cascade}
\bibfield{author}{\bibinfo{person}{Zhaowei Cai} {and} \bibinfo{person}{Nuno
  Vasconcelos}.} \bibinfo{year}{2018}\natexlab{}.
\newblock \showarticletitle{Cascade {R-CNN:} Delving Into High Quality Object
  Detection}. In \bibinfo{booktitle}{\emph{2018 {IEEE} Conference on Computer
  Vision and Pattern Recognition, {CVPR} 2018, Salt Lake City, UT, USA, June
  18-22, 2018}}. \bibinfo{publisher}{{IEEE} Computer Society},
  \bibinfo{pages}{6154--6162}.
\newblock
\urldef\tempurl%
\url{https://doi.org/10.1109/CVPR.2018.00644}
\showDOI{\tempurl}


\bibitem[Carion et~al\mbox{.}(2020)]%
        {carion2020end}
\bibfield{author}{\bibinfo{person}{Nicolas Carion}, \bibinfo{person}{Francisco
  Massa}, \bibinfo{person}{Gabriel Synnaeve}, \bibinfo{person}{Nicolas
  Usunier}, \bibinfo{person}{Alexander Kirillov}, {and} \bibinfo{person}{Sergey
  Zagoruyko}.} \bibinfo{year}{2020}\natexlab{}.
\newblock \showarticletitle{End-to-end object detection with transformers}. In
  \bibinfo{booktitle}{\emph{European conference on computer vision}}. Springer,
  \bibinfo{pages}{213--229}.
\newblock


\bibitem[Caron et~al\mbox{.}(2021)]%
        {caron2021emerging}
\bibfield{author}{\bibinfo{person}{Mathilde Caron}, \bibinfo{person}{Hugo
  Touvron}, \bibinfo{person}{Ishan Misra}, \bibinfo{person}{Hervé Jégou},
  \bibinfo{person}{Julien Mairal}, \bibinfo{person}{Piotr Bojanowski}, {and}
  \bibinfo{person}{Armand Joulin}.} \bibinfo{year}{2021}\natexlab{}.
\newblock \bibinfo{title}{Emerging Properties in Self-Supervised Vision
  Transformers}.
\newblock
\newblock
\showeprint[arxiv]{2104.14294}~[cs.CV]


\bibitem[Chen et~al\mbox{.}(2020)]%
        {pmlr-v119-chen20s}
\bibfield{author}{\bibinfo{person}{Mark Chen}, \bibinfo{person}{Alec Radford},
  \bibinfo{person}{Rewon Child}, \bibinfo{person}{Jeffrey Wu},
  \bibinfo{person}{Heewoo Jun}, \bibinfo{person}{David Luan}, {and}
  \bibinfo{person}{Ilya Sutskever}.} \bibinfo{year}{2020}\natexlab{}.
\newblock \showarticletitle{Generative Pretraining From Pixels}. In
  \bibinfo{booktitle}{\emph{Proceedings of the 37th International Conference on
  Machine Learning, {ICML} 2020, 13-18 July 2020, Virtual Event}}
  \emph{(\bibinfo{series}{Proceedings of Machine Learning Research},
  Vol.~\bibinfo{volume}{119})}. \bibinfo{publisher}{{PMLR}},
  \bibinfo{pages}{1691--1703}.
\newblock
\urldef\tempurl%
\url{http://proceedings.mlr.press/v119/chen20s.html}
\showURL{%
\tempurl}


\bibitem[Chen et~al\mbox{.}(2021)]%
        {chen2021empirical}
\bibfield{author}{\bibinfo{person}{Xinlei Chen}, \bibinfo{person}{Saining Xie},
  {and} \bibinfo{person}{Kaiming He}.} \bibinfo{year}{2021}\natexlab{}.
\newblock \bibinfo{title}{An Empirical Study of Training Self-Supervised Vision
  Transformers}.
\newblock
\newblock
\showeprint[arxiv]{2104.02057}~[cs.CV]


\bibitem[Cui et~al\mbox{.}(2021)]%
        {cui2021document}
\bibfield{author}{\bibinfo{person}{Lei Cui}, \bibinfo{person}{Yiheng Xu},
  \bibinfo{person}{Tengchao Lv}, {and} \bibinfo{person}{Furu Wei}.}
  \bibinfo{year}{2021}\natexlab{}.
\newblock \bibinfo{title}{Document AI: Benchmarks, Models and Applications}.
\newblock
\newblock
\showeprint[arxiv]{2111.08609}~[cs.CL]


\bibitem[Das et~al\mbox{.}(2018)]%
        {Das2018DocumentIC}
\bibfield{author}{\bibinfo{person}{Arindam Das}, \bibinfo{person}{Saikat Roy},
  {and} \bibinfo{person}{Ujjwal Bhattacharya}.}
  \bibinfo{year}{2018}\natexlab{}.
\newblock \showarticletitle{Document Image Classification with Intra-Domain
  Transfer Learning and Stacked Generalization of Deep Convolutional Neural
  Networks}.
\newblock \bibinfo{journal}{\emph{2018 24th International Conference on Pattern
  Recognition (ICPR)}} (\bibinfo{year}{2018}), \bibinfo{pages}{3180--3185}.
\newblock


\bibitem[Dosovitskiy et~al\mbox{.}(2021)]%
        {dosovitskiy2020vit}
\bibfield{author}{\bibinfo{person}{Alexey Dosovitskiy}, \bibinfo{person}{Lucas
  Beyer}, \bibinfo{person}{Alexander Kolesnikov}, \bibinfo{person}{Dirk
  Weissenborn}, \bibinfo{person}{Xiaohua Zhai}, \bibinfo{person}{Thomas
  Unterthiner}, \bibinfo{person}{Mostafa Dehghani}, \bibinfo{person}{Matthias
  Minderer}, \bibinfo{person}{Georg Heigold}, \bibinfo{person}{Sylvain Gelly},
  \bibinfo{person}{Jakob Uszkoreit}, {and} \bibinfo{person}{Neil Houlsby}.}
  \bibinfo{year}{2021}\natexlab{}.
\newblock \showarticletitle{An Image is Worth 16x16 Words: Transformers for
  Image Recognition at Scale}.
\newblock \bibinfo{journal}{\emph{ICLR}} (\bibinfo{year}{2021}).
\newblock


\bibitem[El-Nouby et~al\mbox{.}(2021a)]%
        {elnouby2021xcit}
\bibfield{author}{\bibinfo{person}{Alaaeldin El-Nouby}, \bibinfo{person}{Hugo
  Touvron}, \bibinfo{person}{Mathilde Caron}, \bibinfo{person}{Piotr
  Bojanowski}, \bibinfo{person}{Matthijs Douze}, \bibinfo{person}{Armand
  Joulin}, \bibinfo{person}{Ivan Laptev}, \bibinfo{person}{Natalia Neverova},
  \bibinfo{person}{Gabriel Synnaeve}, \bibinfo{person}{Jakob Verbeek}, {and}
  \bibinfo{person}{Hervé Jegou}.} \bibinfo{year}{2021}\natexlab{a}.
\newblock \bibinfo{title}{XCiT: Cross-Covariance Image Transformers}.
\newblock
\newblock
\showeprint[arxiv]{2106.09681}~[cs.CV]


\bibitem[El-Nouby et~al\mbox{.}(2021b)]%
        {ElNouby2021XCiTCI}
\bibfield{author}{\bibinfo{person}{Alaaeldin El-Nouby}, \bibinfo{person}{Hugo
  Touvron}, \bibinfo{person}{Mathilde Caron}, \bibinfo{person}{Piotr
  Bojanowski}, \bibinfo{person}{Matthijs Douze}, \bibinfo{person}{Armand
  Joulin}, \bibinfo{person}{Ivan Laptev}, \bibinfo{person}{Natalia Neverova},
  \bibinfo{person}{Gabriel Synnaeve}, \bibinfo{person}{Jakob Verbeek}, {and}
  \bibinfo{person}{Herv{\'e} J{\'e}gou}.} \bibinfo{year}{2021}\natexlab{b}.
\newblock \showarticletitle{XCiT: Cross-Covariance Image Transformers}.
\newblock \bibinfo{journal}{\emph{ArXiv}}  \bibinfo{volume}{abs/2106.09681}
  (\bibinfo{year}{2021}).
\newblock


\bibitem[Gao et~al\mbox{.}(2019)]%
        {8978120}
\bibfield{author}{\bibinfo{person}{Liangcai Gao}, \bibinfo{person}{Yilun
  Huang}, \bibinfo{person}{Hervé Déjean}, \bibinfo{person}{Jean-Luc Meunier},
  \bibinfo{person}{Qinqin Yan}, \bibinfo{person}{Yu Fang},
  \bibinfo{person}{Florian Kleber}, {and} \bibinfo{person}{Eva Lang}.}
  \bibinfo{year}{2019}\natexlab{}.
\newblock \showarticletitle{ICDAR 2019 Competition on Table Detection and
  Recognition (cTDaR)}. In \bibinfo{booktitle}{\emph{2019 International
  Conference on Document Analysis and Recognition (ICDAR)}}.
  \bibinfo{pages}{1510--1515}.
\newblock
\urldef\tempurl%
\url{https://doi.org/10.1109/ICDAR.2019.00243}
\showDOI{\tempurl}


\bibitem[Harley et~al\mbox{.}(2015)]%
        {harley2015icdar}
\bibfield{author}{\bibinfo{person}{Adam~W Harley}, \bibinfo{person}{Alex
  Ufkes}, {and} \bibinfo{person}{Konstantinos~G Derpanis}.}
  \bibinfo{year}{2015}\natexlab{}.
\newblock \showarticletitle{Evaluation of Deep Convolutional Nets for Document
  Image Classification and Retrieval}. In
  \bibinfo{booktitle}{\emph{International Conference on Document Analysis and
  Recognition ({ICDAR})}}.
\newblock


\bibitem[He et~al\mbox{.}(2021)]%
        {he2021masked}
\bibfield{author}{\bibinfo{person}{Kaiming He}, \bibinfo{person}{Xinlei Chen},
  \bibinfo{person}{Saining Xie}, \bibinfo{person}{Yanghao Li},
  \bibinfo{person}{Piotr Dollár}, {and} \bibinfo{person}{Ross Girshick}.}
  \bibinfo{year}{2021}\natexlab{}.
\newblock \bibinfo{title}{Masked Autoencoders Are Scalable Vision Learners}.
\newblock
\newblock
\showeprint[arxiv]{2111.06377}~[cs.CV]


\bibitem[He et~al\mbox{.}(2017)]%
        {he2017mask}
\bibfield{author}{\bibinfo{person}{Kaiming He}, \bibinfo{person}{Georgia
  Gkioxari}, \bibinfo{person}{Piotr Doll{\'{a}}r}, {and}
  \bibinfo{person}{Ross~B. Girshick}.} \bibinfo{year}{2017}\natexlab{}.
\newblock \showarticletitle{Mask {R-CNN}}. In \bibinfo{booktitle}{\emph{{IEEE}
  International Conference on Computer Vision, {ICCV} 2017, Venice, Italy,
  October 22-29, 2017}}. \bibinfo{publisher}{{IEEE} Computer Society},
  \bibinfo{pages}{2980--2988}.
\newblock
\urldef\tempurl%
\url{https://doi.org/10.1109/ICCV.2017.322}
\showDOI{\tempurl}


\bibitem[Hong et~al\mbox{.}(2021)]%
        {hong2021bros}
\bibfield{author}{\bibinfo{person}{Teakgyu Hong}, \bibinfo{person}{DongHyun
  Kim}, \bibinfo{person}{Mingi Ji}, \bibinfo{person}{Wonseok Hwang},
  \bibinfo{person}{Daehyun Nam}, {and} \bibinfo{person}{Sungrae Park}.}
  \bibinfo{year}{2021}\natexlab{}.
\newblock \bibinfo{title}{BROS: A Pre-trained Language Model for Understanding
  Texts in Document}.
\newblock
\newblock
\urldef\tempurl%
\url{https://openreview.net/forum?id=punMXQEsPr0}
\showURL{%
\tempurl}


\bibitem[Huang et~al\mbox{.}(2016)]%
        {Huang2016DeepNW}
\bibfield{author}{\bibinfo{person}{Gao Huang}, \bibinfo{person}{Yu Sun},
  \bibinfo{person}{Zhuang Liu}, \bibinfo{person}{Daniel Sedra}, {and}
  \bibinfo{person}{Kilian~Q. Weinberger}.} \bibinfo{year}{2016}\natexlab{}.
\newblock \showarticletitle{Deep Networks with Stochastic Depth}. In
  \bibinfo{booktitle}{\emph{ECCV}}.
\newblock


\bibitem[Huang et~al\mbox{.}(2022)]%
        {huang2022layoutlmv3}
\bibfield{author}{\bibinfo{person}{Yupan Huang}, \bibinfo{person}{Tengchao Lv},
  \bibinfo{person}{Lei Cui}, \bibinfo{person}{Yutong Lu}, {and}
  \bibinfo{person}{Furu Wei}.} \bibinfo{year}{2022}\natexlab{}.
\newblock \showarticletitle{{LayoutLMv3:} Pre-training for Document AI with
  Unified Text and Image Masking}. In \bibinfo{booktitle}{\emph{{MM} '22: The
  30th {ACM} International Conference on Multimedia, Lisbon,Portugal, October
  10-14, 2022}}.
\newblock


\bibitem[Jaume et~al\mbox{.}(2019)]%
        {Jaume2019FUNSDAD}
\bibfield{author}{\bibinfo{person}{Guillaume Jaume},
  \bibinfo{person}{Hazim~Kemal Ekenel}, {and} \bibinfo{person}{Jean-Philippe
  Thiran}.} \bibinfo{year}{2019}\natexlab{}.
\newblock \showarticletitle{FUNSD: A Dataset for Form Understanding in Noisy
  Scanned Documents}.
\newblock \bibinfo{journal}{\emph{2019 International Conference on Document
  Analysis and Recognition Workshops (ICDARW)}}  \bibinfo{volume}{2}
  (\bibinfo{year}{2019}), \bibinfo{pages}{1--6}.
\newblock


\bibitem[Kingma and Ba(2015)]%
        {Kingma2015AdamAM}
\bibfield{author}{\bibinfo{person}{Diederik~P. Kingma} {and}
  \bibinfo{person}{Jimmy Ba}.} \bibinfo{year}{2015}\natexlab{}.
\newblock \showarticletitle{Adam: {A} Method for Stochastic Optimization}. In
  \bibinfo{booktitle}{\emph{3rd International Conference on Learning
  Representations, {ICLR} 2015, San Diego, CA, USA, May 7-9, 2015, Conference
  Track Proceedings}}, \bibfield{editor}{\bibinfo{person}{Yoshua Bengio} {and}
  \bibinfo{person}{Yann LeCun}} (Eds.).
\newblock
\urldef\tempurl%
\url{http://arxiv.org/abs/1412.6980}
\showURL{%
\tempurl}


\bibitem[Lewis et~al\mbox{.}(2006)]%
        {Lewis:2006:BTC:1148170.1148307}
\bibfield{author}{\bibinfo{person}{D. Lewis}, \bibinfo{person}{G. Agam},
  \bibinfo{person}{S. Argamon}, \bibinfo{person}{O. Frieder},
  \bibinfo{person}{D. Grossman}, {and} \bibinfo{person}{J. Heard}.}
  \bibinfo{year}{2006}\natexlab{}.
\newblock \showarticletitle{Building a Test Collection for Complex Document
  Information Processing}. In \bibinfo{booktitle}{\emph{Proceedings of the 29th
  Annual International ACM SIGIR Conference on Research and Development in
  Information Retrieval}} (Seattle, Washington, USA)
  \emph{(\bibinfo{series}{SIGIR '06})}. \bibinfo{publisher}{ACM},
  \bibinfo{address}{New York, NY, USA}, \bibinfo{pages}{665--666}.
\newblock
\showISBNx{1-59593-369-7}
\urldef\tempurl%
\url{https://doi.org/10.1145/1148170.1148307}
\showDOI{\tempurl}


\bibitem[Li et~al\mbox{.}(2021a)]%
        {li2021structurallm}
\bibfield{author}{\bibinfo{person}{Chenliang Li}, \bibinfo{person}{Bin Bi},
  \bibinfo{person}{Ming Yan}, \bibinfo{person}{Wei Wang},
  \bibinfo{person}{Songfang Huang}, \bibinfo{person}{Fei Huang}, {and}
  \bibinfo{person}{Luo Si}.} \bibinfo{year}{2021}\natexlab{a}.
\newblock \showarticletitle{{S}tructural{LM}: Structural Pre-training for Form
  Understanding}. In \bibinfo{booktitle}{\emph{Proceedings of the 59th Annual
  Meeting of the Association for Computational Linguistics and the 11th
  International Joint Conference on Natural Language Processing (Volume 1: Long
  Papers)}}. \bibinfo{publisher}{Association for Computational Linguistics},
  \bibinfo{address}{Online}, \bibinfo{pages}{6309--6318}.
\newblock
\urldef\tempurl%
\url{https://doi.org/10.18653/v1/2021.acl-long.493}
\showDOI{\tempurl}


\bibitem[Li et~al\mbox{.}(2020a)]%
        {li-etal-2020-tablebank}
\bibfield{author}{\bibinfo{person}{Minghao Li}, \bibinfo{person}{Lei Cui},
  \bibinfo{person}{Shaohan Huang}, \bibinfo{person}{Furu Wei},
  \bibinfo{person}{Ming Zhou}, {and} \bibinfo{person}{Zhoujun Li}.}
  \bibinfo{year}{2020}\natexlab{a}.
\newblock \showarticletitle{{T}able{B}ank: Table Benchmark for Image-based
  Table Detection and Recognition}. In \bibinfo{booktitle}{\emph{Proceedings of
  the 12th Language Resources and Evaluation Conference}}.
  \bibinfo{publisher}{European Language Resources Association},
  \bibinfo{address}{Marseille, France}, \bibinfo{pages}{1918--1925}.
\newblock
\showISBNx{979-10-95546-34-4}
\urldef\tempurl%
\url{https://aclanthology.org/2020.lrec-1.236}
\showURL{%
\tempurl}


\bibitem[Li et~al\mbox{.}(2020b)]%
        {li-etal-2020-docbank}
\bibfield{author}{\bibinfo{person}{Minghao Li}, \bibinfo{person}{Yiheng Xu},
  \bibinfo{person}{Lei Cui}, \bibinfo{person}{Shaohan Huang},
  \bibinfo{person}{Furu Wei}, \bibinfo{person}{Zhoujun Li}, {and}
  \bibinfo{person}{Ming Zhou}.} \bibinfo{year}{2020}\natexlab{b}.
\newblock \showarticletitle{{D}oc{B}ank: A Benchmark Dataset for Document
  Layout Analysis}. In \bibinfo{booktitle}{\emph{Proceedings of the 28th
  International Conference on Computational Linguistics}}.
  \bibinfo{publisher}{International Committee on Computational Linguistics},
  \bibinfo{address}{Barcelona, Spain (Online)}, \bibinfo{pages}{949--960}.
\newblock
\urldef\tempurl%
\url{https://doi.org/10.18653/v1/2020.coling-main.82}
\showDOI{\tempurl}


\bibitem[Li et~al\mbox{.}(2021b)]%
        {li2021selfdoc}
\bibfield{author}{\bibinfo{person}{Peizhao Li}, \bibinfo{person}{Jiuxiang Gu},
  \bibinfo{person}{Jason Kuen}, \bibinfo{person}{Vlad~I. Morariu},
  \bibinfo{person}{Handong Zhao}, \bibinfo{person}{Rajiv Jain},
  \bibinfo{person}{Varun Manjunatha}, {and} \bibinfo{person}{Hongfu Liu}.}
  \bibinfo{year}{2021}\natexlab{b}.
\newblock \bibinfo{title}{SelfDoc: Self-Supervised Document Representation
  Learning}.
\newblock
\newblock
\showeprint[arxiv]{2106.03331}~[cs.CV]


\bibitem[Li et~al\mbox{.}(2021c)]%
        {Li2021BenchmarkingDT}
\bibfield{author}{\bibinfo{person}{Yanghao Li}, \bibinfo{person}{Saining Xie},
  \bibinfo{person}{Xinlei Chen}, \bibinfo{person}{Piotr Doll{\'a}r},
  \bibinfo{person}{Kaiming He}, {and} \bibinfo{person}{Ross~B. Girshick}.}
  \bibinfo{year}{2021}\natexlab{c}.
\newblock \showarticletitle{Benchmarking Detection Transfer Learning with
  Vision Transformers}.
\newblock \bibinfo{journal}{\emph{ArXiv}}  \bibinfo{volume}{abs/2111.11429}
  (\bibinfo{year}{2021}).
\newblock


\bibitem[Liao et~al\mbox{.}(2022)]%
        {liao2022realtime}
\bibfield{author}{\bibinfo{person}{Minghui Liao}, \bibinfo{person}{Zhisheng
  Zou}, \bibinfo{person}{Zhaoyi Wan}, \bibinfo{person}{Cong Yao}, {and}
  \bibinfo{person}{Xiang Bai}.} \bibinfo{year}{2022}\natexlab{}.
\newblock \bibinfo{title}{Real-Time Scene Text Detection with Differentiable
  Binarization and Adaptive Scale Fusion}.
\newblock
\newblock
\showeprint[arxiv]{2202.10304}~[cs.CV]


\bibitem[Liu et~al\mbox{.}(2021)]%
        {liu2021swin}
\bibfield{author}{\bibinfo{person}{Ze Liu}, \bibinfo{person}{Yutong Lin},
  \bibinfo{person}{Yue Cao}, \bibinfo{person}{Han Hu}, \bibinfo{person}{Yixuan
  Wei}, \bibinfo{person}{Zheng Zhang}, \bibinfo{person}{Stephen Lin}, {and}
  \bibinfo{person}{Baining Guo}.} \bibinfo{year}{2021}\natexlab{}.
\newblock \bibinfo{title}{Swin Transformer: Hierarchical Vision Transformer
  using Shifted Windows}.
\newblock
\newblock
\showeprint[arxiv]{2103.14030}~[cs.CV]


\bibitem[Powalski et~al\mbox{.}(2021)]%
        {powalski2021going}
\bibfield{author}{\bibinfo{person}{Rafał Powalski}, \bibinfo{person}{Łukasz
  Borchmann}, \bibinfo{person}{Dawid Jurkiewicz}, \bibinfo{person}{Tomasz
  Dwojak}, \bibinfo{person}{Michał Pietruszka}, {and}
  \bibinfo{person}{Gabriela Pałka}.} \bibinfo{year}{2021}\natexlab{}.
\newblock \bibinfo{title}{Going Full-TILT Boogie on Document Understanding with
  Text-Image-Layout Transformer}.
\newblock
\newblock
\showeprint[arxiv]{2102.09550}~[cs.CL]


\bibitem[Pramanik et~al\mbox{.}(2020)]%
        {pramanik2020multimodal}
\bibfield{author}{\bibinfo{person}{Subhojeet Pramanik},
  \bibinfo{person}{Shashank Mujumdar}, {and} \bibinfo{person}{Hima Patel}.}
  \bibinfo{year}{2020}\natexlab{}.
\newblock \bibinfo{title}{Towards a Multi-modal, Multi-task Learning based
  Pre-training Framework for Document Representation Learning}.
\newblock
\newblock
\showeprint[arxiv]{2009.14457}~[cs.CL]


\bibitem[Ramesh et~al\mbox{.}(2021)]%
        {ramesh2021zeroshot}
\bibfield{author}{\bibinfo{person}{Aditya Ramesh}, \bibinfo{person}{Mikhail
  Pavlov}, \bibinfo{person}{Gabriel Goh}, \bibinfo{person}{Scott Gray},
  \bibinfo{person}{Chelsea Voss}, \bibinfo{person}{Alec Radford},
  \bibinfo{person}{Mark Chen}, {and} \bibinfo{person}{Ilya Sutskever}.}
  \bibinfo{year}{2021}\natexlab{}.
\newblock \bibinfo{title}{Zero-Shot Text-to-Image Generation}.
\newblock
\newblock
\showeprint[arxiv]{2102.12092}~[cs.CV]


\bibitem[Sarkhel and Nandi(2019)]%
        {ijcai2019-466}
\bibfield{author}{\bibinfo{person}{Ritesh Sarkhel} {and} \bibinfo{person}{Arnab
  Nandi}.} \bibinfo{year}{2019}\natexlab{}.
\newblock \showarticletitle{Deterministic Routing between Layout Abstractions
  for Multi-Scale Classification of Visually Rich Documents}. In
  \bibinfo{booktitle}{\emph{Proceedings of the Twenty-Eighth International
  Joint Conference on Artificial Intelligence, {IJCAI} 2019, Macao, China,
  August 10-16, 2019}}, \bibfield{editor}{\bibinfo{person}{Sarit Kraus}} (Ed.).
  \bibinfo{publisher}{ijcai.org}, \bibinfo{pages}{3360--3366}.
\newblock
\urldef\tempurl%
\url{https://doi.org/10.24963/ijcai.2019/466}
\showDOI{\tempurl}


\bibitem[Touvron et~al\mbox{.}(2021)]%
        {touvron2020deit}
\bibfield{author}{\bibinfo{person}{Hugo Touvron}, \bibinfo{person}{Matthieu
  Cord}, \bibinfo{person}{Matthijs Douze}, \bibinfo{person}{Francisco Massa},
  \bibinfo{person}{Alexandre Sablayrolles}, {and} \bibinfo{person}{Herv{\'e}
  J{\'e}gou}.} \bibinfo{year}{2021}\natexlab{}.
\newblock \showarticletitle{Training data-efficient image transformers \&
  distillation through attention}. In \bibinfo{booktitle}{\emph{International
  Conference on Machine Learning}}. PMLR, \bibinfo{pages}{10347--10357}.
\newblock


\bibitem[Vaswani et~al\mbox{.}(2017)]%
        {vaswani2017attention}
\bibfield{author}{\bibinfo{person}{Ashish Vaswani}, \bibinfo{person}{Noam
  Shazeer}, \bibinfo{person}{Niki Parmar}, \bibinfo{person}{Jakob Uszkoreit},
  \bibinfo{person}{Llion Jones}, \bibinfo{person}{Aidan~N. Gomez},
  \bibinfo{person}{Lukasz Kaiser}, {and} \bibinfo{person}{Illia Polosukhin}.}
  \bibinfo{year}{2017}\natexlab{}.
\newblock \showarticletitle{Attention is All you Need}. In
  \bibinfo{booktitle}{\emph{Advances in Neural Information Processing Systems
  30: Annual Conference on Neural Information Processing Systems 2017, December
  4-9, 2017, Long Beach, CA, {USA}}},
  \bibfield{editor}{\bibinfo{person}{Isabelle Guyon}, \bibinfo{person}{Ulrike
  von Luxburg}, \bibinfo{person}{Samy Bengio}, \bibinfo{person}{Hanna~M.
  Wallach}, \bibinfo{person}{Rob Fergus}, \bibinfo{person}{S.~V.~N.
  Vishwanathan}, {and} \bibinfo{person}{Roman Garnett}} (Eds.).
  \bibinfo{pages}{5998--6008}.
\newblock
\urldef\tempurl%
\url{https://proceedings.neurips.cc/paper/2017/hash/3f5ee243547dee91fbd053c1c4a845aa-Abstract.html}
\showURL{%
\tempurl}


\bibitem[Wu et~al\mbox{.}(2021)]%
        {wu2021lampret}
\bibfield{author}{\bibinfo{person}{Te-Lin Wu}, \bibinfo{person}{Cheng Li},
  \bibinfo{person}{Mingyang Zhang}, \bibinfo{person}{Tao Chen},
  \bibinfo{person}{Spurthi~Amba Hombaiah}, {and} \bibinfo{person}{Michael
  Bendersky}.} \bibinfo{year}{2021}\natexlab{}.
\newblock \bibinfo{title}{LAMPRET: Layout-Aware Multimodal PreTraining for
  Document Understanding}.
\newblock
\newblock
\showeprint[arxiv]{2104.08405}~[cs.CL]


\bibitem[Wu et~al\mbox{.}(2019)]%
        {wu2019detectron2}
\bibfield{author}{\bibinfo{person}{Yuxin Wu}, \bibinfo{person}{Alexander
  Kirillov}, \bibinfo{person}{Francisco Massa}, \bibinfo{person}{Wan-Yen Lo},
  {and} \bibinfo{person}{Ross Girshick}.} \bibinfo{year}{2019}\natexlab{}.
\newblock \bibinfo{title}{Detectron2}.
\newblock
  \bibinfo{howpublished}{\url{https://github.com/facebookresearch/detectron2}}.
\newblock


\bibitem[Xie et~al\mbox{.}(2017)]%
        {Xie2016AggregatedRT}
\bibfield{author}{\bibinfo{person}{Saining Xie}, \bibinfo{person}{Ross~B.
  Girshick}, \bibinfo{person}{Piotr Doll{\'{a}}r}, \bibinfo{person}{Zhuowen
  Tu}, {and} \bibinfo{person}{Kaiming He}.} \bibinfo{year}{2017}\natexlab{}.
\newblock \showarticletitle{Aggregated Residual Transformations for Deep Neural
  Networks}. In \bibinfo{booktitle}{\emph{2017 {IEEE} Conference on Computer
  Vision and Pattern Recognition, {CVPR} 2017, Honolulu, HI, USA, July 21-26,
  2017}}. \bibinfo{publisher}{{IEEE} Computer Society},
  \bibinfo{pages}{5987--5995}.
\newblock
\urldef\tempurl%
\url{https://doi.org/10.1109/CVPR.2017.634}
\showDOI{\tempurl}


\bibitem[Xu et~al\mbox{.}(2020)]%
        {10.1145/3394486.3403172}
\bibfield{author}{\bibinfo{person}{Yiheng Xu}, \bibinfo{person}{Minghao Li},
  \bibinfo{person}{Lei Cui}, \bibinfo{person}{Shaohan Huang},
  \bibinfo{person}{Furu Wei}, {and} \bibinfo{person}{Ming Zhou}.}
  \bibinfo{year}{2020}\natexlab{}.
\newblock \showarticletitle{LayoutLM: Pre-training of Text and Layout for
  Document Image Understanding}. In \bibinfo{booktitle}{\emph{{KDD} '20: The
  26th {ACM} {SIGKDD} Conference on Knowledge Discovery and Data Mining,
  Virtual Event, CA, USA, August 23-27, 2020}},
  \bibfield{editor}{\bibinfo{person}{Rajesh Gupta}, \bibinfo{person}{Yan Liu},
  \bibinfo{person}{Jiliang Tang}, {and} \bibinfo{person}{B.~Aditya Prakash}}
  (Eds.). \bibinfo{publisher}{{ACM}}, \bibinfo{pages}{1192--1200}.
\newblock
\urldef\tempurl%
\url{https://dl.acm.org/doi/10.1145/3394486.3403172}
\showURL{%
\tempurl}


\bibitem[Xu et~al\mbox{.}(2021a)]%
        {xu2021layoutxlm}
\bibfield{author}{\bibinfo{person}{Yiheng Xu}, \bibinfo{person}{Tengchao Lv},
  \bibinfo{person}{Lei Cui}, \bibinfo{person}{Guoxin Wang},
  \bibinfo{person}{Yijuan Lu}, \bibinfo{person}{Dinei Florencio},
  \bibinfo{person}{Cha Zhang}, {and} \bibinfo{person}{Furu Wei}.}
  \bibinfo{year}{2021}\natexlab{a}.
\newblock \bibinfo{title}{LayoutXLM: Multimodal Pre-training for Multilingual
  Visually-rich Document Understanding}.
\newblock
\newblock
\showeprint[arxiv]{2104.08836}~[cs.CL]


\bibitem[Xu et~al\mbox{.}(2022)]%
        {xu-etal-2022-xfund}
\bibfield{author}{\bibinfo{person}{Yiheng Xu}, \bibinfo{person}{Tengchao Lv},
  \bibinfo{person}{Lei Cui}, \bibinfo{person}{Guoxin Wang},
  \bibinfo{person}{Yijuan Lu}, \bibinfo{person}{Dinei Florencio},
  \bibinfo{person}{Cha Zhang}, {and} \bibinfo{person}{Furu Wei}.}
  \bibinfo{year}{2022}\natexlab{}.
\newblock \showarticletitle{{XFUND}: A Benchmark Dataset for Multilingual
  Visually Rich Form Understanding}. In \bibinfo{booktitle}{\emph{Findings of
  the Association for Computational Linguistics: ACL 2022}}.
  \bibinfo{publisher}{Association for Computational Linguistics},
  \bibinfo{address}{Dublin, Ireland}, \bibinfo{pages}{3214--3224}.
\newblock
\urldef\tempurl%
\url{https://doi.org/10.18653/v1/2022.findings-acl.253}
\showDOI{\tempurl}


\bibitem[Xu et~al\mbox{.}(2021b)]%
        {xu2021layoutlmv2}
\bibfield{author}{\bibinfo{person}{Yang Xu}, \bibinfo{person}{Yiheng Xu},
  \bibinfo{person}{Tengchao Lv}, \bibinfo{person}{Lei Cui},
  \bibinfo{person}{Furu Wei}, \bibinfo{person}{Guoxin Wang},
  \bibinfo{person}{Yijuan Lu}, \bibinfo{person}{Dinei Florencio},
  \bibinfo{person}{Cha Zhang}, \bibinfo{person}{Wanxiang Che},
  \bibinfo{person}{Min Zhang}, {and} \bibinfo{person}{Lidong Zhou}.}
  \bibinfo{year}{2021}\natexlab{b}.
\newblock \showarticletitle{{L}ayout{LM}v2: Multi-modal Pre-training for
  Visually-rich Document Understanding}. In
  \bibinfo{booktitle}{\emph{Proceedings of the 59th Annual Meeting of the
  Association for Computational Linguistics and the 11th International Joint
  Conference on Natural Language Processing (Volume 1: Long Papers)}}.
  \bibinfo{publisher}{Association for Computational Linguistics},
  \bibinfo{address}{Online}, \bibinfo{pages}{2579--2591}.
\newblock
\urldef\tempurl%
\url{https://doi.org/10.18653/v1/2021.acl-long.201}
\showDOI{\tempurl}


\bibitem[Zhong et~al\mbox{.}(2020)]%
        {zhong2020imagebased}
\bibfield{author}{\bibinfo{person}{Xu Zhong}, \bibinfo{person}{Elaheh
  ShafieiBavani}, {and} \bibinfo{person}{Antonio~Jimeno Yepes}.}
  \bibinfo{year}{2020}\natexlab{}.
\newblock \bibinfo{title}{Image-based table recognition: data, model, and
  evaluation}.
\newblock
\newblock
\showeprint[arxiv]{1911.10683}~[cs.CV]


\bibitem[Zhong et~al\mbox{.}(2019)]%
        {zhong2019publaynet}
\bibfield{author}{\bibinfo{person}{Xu Zhong}, \bibinfo{person}{Jianbin Tang},
  {and} \bibinfo{person}{Antonio~Jimeno Yepes}.}
  \bibinfo{year}{2019}\natexlab{}.
\newblock \showarticletitle{PubLayNet: largest dataset ever for document layout
  analysis}. In \bibinfo{booktitle}{\emph{2019 International Conference on
  Document Analysis and Recognition (ICDAR)}}. IEEE,
  \bibinfo{pages}{1015--1022}.
\newblock
\showISSN{1520-5363}
\urldef\tempurl%
\url{https://doi.org/10.1109/ICDAR.2019.00166}
\showDOI{\tempurl}


\bibitem[Zhou et~al\mbox{.}(2021)]%
        {zhou2021ibot}
\bibfield{author}{\bibinfo{person}{Jinghao Zhou}, \bibinfo{person}{Chen Wei},
  \bibinfo{person}{Huiyu Wang}, \bibinfo{person}{Wei Shen},
  \bibinfo{person}{Cihang Xie}, \bibinfo{person}{Alan Yuille}, {and}
  \bibinfo{person}{Tao Kong}.} \bibinfo{year}{2021}\natexlab{}.
\newblock \bibinfo{title}{iBOT: Image BERT Pre-Training with Online Tokenizer}.
\newblock
\newblock
\showeprint[arxiv]{2111.07832}~[cs.CV]


\bibitem[Łukasz Garncarek et~al\mbox{.}(2021)]%
        {garncarek2021lambert}
\bibfield{author}{\bibinfo{person}{Łukasz Garncarek}, \bibinfo{person}{Rafał
  Powalski}, \bibinfo{person}{Tomasz Stanisławek}, \bibinfo{person}{Bartosz
  Topolski}, \bibinfo{person}{Piotr Halama}, \bibinfo{person}{Michał Turski},
  {and} \bibinfo{person}{Filip Graliński}.} \bibinfo{year}{2021}\natexlab{}.
\newblock \bibinfo{title}{LAMBERT: Layout-Aware (Language) Modeling for
  information extraction}.
\newblock
\newblock
\showeprint[arxiv]{2002.08087}~[cs.CL]


\end{thebibliography}

\end{document}